\documentclass[10pt]{article} %
\usepackage[accepted]{TMLR/tmlr}

\usepackage{hyperref}
\usepackage{url}

\title{FuseLIP:\\Multimodal Embeddings via Early Fusion of Discrete Tokens}

\author{\name Christian Schlarmann \email christian.schlarmann@uni-tuebingen.de \\
      \addr Tübingen AI Center, University of Tübingen
      \AND
      \name Francesco Croce \email francesco.croce@aalto.fi \\
      \addr ELLIS Institute Finland, Aalto University
      \AND
      \name Nicolas Flammarion \email  nicolas.flammarion@epfl.ch \\
      \addr EPFL
      \AND Matthias Hein \email matthias.hein@uni-tuebingen.de \\
      \addr Tübingen AI Center, University of Tübingen
      }

\def\month{08}  %
\def\year{2026} %
\def\openreview{\url{https://openreview.net/forum?id=yq9je6kLC6}} %

\usepackage[dvipsnames]{xcolor}         %
\usepackage{booktabs}
\usepackage{multirow}
\usepackage{pifont}
\usepackage{listings}
\usepackage{caption}
\usepackage{bbm}
\usepackage{arydshln}
\usepackage{listings}
\usepackage{placeins}
\usepackage{nicefrac}
\usepackage{xurl}
\usepackage{enumitem}
\usepackage{tabularx}
\usepackage{array}
\usepackage{nicefrac}
\usepackage{makecell}
\usepackage{xspace}
\usepackage{graphbox}
\usepackage{pifont}
\usepackage{enumitem}
\usepackage{colortbl}
\usepackage{multirow}
\usepackage{diagbox}
\usepackage{array}
\usepackage{wrapfig}
\usepackage{subcaption}
\usepackage{footmisc}
\usepackage{color}
\usepackage{stfloats}
\usepackage{slashbox}
\usepackage{diagbox}
\usepackage{float}
\usepackage{mathrsfs}
\usepackage{apptools}
\usepackage{bbm}
\usepackage{soul}
\usepackage[export]{adjustbox}
\usepackage{hyperref}
\hypersetup{
    colorlinks=true,
    citecolor=black,
    linkcolor=blue,
    urlcolor=blue
}

\usepackage{amsmath,amsfonts,bm}

\def\eqref#1{(\ref{#1})}

\def\1{\bm{1}}

\def\vtheta{{\bm{\theta}}}

\def\vc{{\bm{c}}}
\def\vd{{\bm{d}}}
\def\ve{{\bm{e}}}

\def\vi{{\bm{i}}}

\def\vt{{\bm{t}}}

\def\vz{{\bm{z}}}

\DeclareMathAlphabet{\mathsfit}{\encodingdefault}{\sfdefault}{m}{sl}
\SetMathAlphabet{\mathsfit}{bold}{\encodingdefault}{\sfdefault}{bx}{n}

\def\R{\mathbb{R}}

\def\R{\mathbb{R}}

\def\L{\mathcal{L}}

\newcommand{\norm}[1]{\left\|#1\right\|}

\definecolor{brightpurple}{RGB}{160,0,255}

\newcommand{\myparagraph}[1]{\noindent\textbf{#1}}

\newcolumntype{C}[1]{>{\centering\arraybackslash}p{#1}}
\newcolumntype{L}[1]{>{\raggedright\arraybackslash}p{#1}}
\newcolumntype{R}[1]{>{\raggedleft\arraybackslash}p{#1}}
\newlength\newl
\newlength\newlc

\newlength\colwidth
\newlength\figwidth

\newcommand{\clip}{CLIP\xspace}
\newcommand{\siglip}{SigLIP\xspace}
\newcommand{\siglipsf}{SigLIP\textsubscript{\textit{SF}}\xspace}
\newcommand{\siglipmlf}{SigLIP\textsubscript{\textit{MLF}}\xspace}
\newcommand{\siglipSsf}{SigLIP-S\textsubscript{\textit{SF}}\xspace}
\newcommand{\siglipSmlf}{SigLIP-S\textsubscript{\textit{MLF}}\xspace}
\newcommand{\siglipBsf}{SigLIP-B\textsubscript{\textit{SF}}\xspace}
\newcommand{\siglipBmlf}{SigLIP-B\textsubscript{\textit{MLF}}\xspace}
\newcommand{\imnet}{ImageNet\xspace}

\newcommand{\vits}{ViT-S/16\xspace}
\newcommand{\vitb}{ViT-B/16\xspace}
\newcommand{\titok}{TiTok\xspace}
\newcommand{\vlmtovec}{VLM2Vec\xspace}
\newcommand{\magiclens}{MagicLens\xspace}

\newcommand{\flava}{FLAVA\xspace}
\newcommand{\ours}{FuseLIP\xspace}
\newcommand{\oursS}{\ours-S\xspace}
\newcommand{\oursB}{\ours-B\xspace}

\newcommand{\cciii}{CC3M\xspace}
\newcommand{\ccxii}{CC12M\xspace}
\newcommand{\cciiivqa}{CC3M-VQA\xspace}
\newcommand{\cciiiaug}{CC3M-TGIT\xspace}
\newcommand{\ccxiiaug}{CC12M-TGIT\xspace}
\newcommand{\cciiimm}{CC3M+MM\xspace}
\newcommand{\ccxiimm}{CC12M+MM\xspace}
\newcommand{\mmdata}{MM\xspace}
\newcommand{\hqedit}{HQ-Edit\xspace}
\newcommand{\vig}{Visual Genome\xspace}
\newcommand{\vgcrop}{VG-Crop\xspace}
\newcommand{\vgvqa}{VG-VQA\xspace}
\newcommand{\oicrop}{OI-Crop\xspace}
\newcommand{\oipos}{OI-Pos\xspace}
\newcommand{\scp}{SugarCrepe\xspace}
\newcommand{\zs}{$^{\star}$}
\newcommand{\cmark}{\textcolor{ForestGreen}{\ding{51}}}  %
\newcommand{\xmark}{\textcolor{red}{\ding{55}}}  %

\newcommand{\chst}[1]{{\color{black}#1}}

\newcommand{\revXRX}[1]{{#1}}
\newcommand{\revHYHX}[1]{{#1}}
\newcommand{\revMAoJ}[1]{{#1}}

\definecolor{myBlue}{rgb}{.93,.93,1.}
\newcolumntype{a}{>{\columncolor{myBlue}}c}

\renewcommand{\cite}{\citep}  %

\begin{document}

\maketitle

\begin{abstract}
Contrastive language-image pre-training aligns features of text-image pairs in a common latent space via distinct encoders for each modality. While this approach achieves impressive performance in several zero-shot tasks, it cannot natively handle multimodal inputs, i.e., encoding image and text into a single feature vector. As a remedy, it is common practice to use additional modules to merge the features extracted by unimodal encoders. In this work, we present FuseLIP, a new architecture for multimodal embedding. Leveraging recent progress in discrete image tokenizers, we propose to use a single transformer model operating on a unified vocabulary of text and image tokens. This early fusion approach allows the different modalities to interact at each depth of encoding and obtain richer representations compared to common late fusion. We collect new datasets for multimodal pre-training and evaluation, designing challenging tasks for multimodal encoders. We show that FuseLIP outperforms late fusion approaches in several multimodal and unimodal embedding tasks. 
The code and datasets are released at \url{https://github.com/chs20/fuselip}.
\end{abstract}

\section{Introduction}

Contrastive language-image pre-training (\clip) is a fundamental approach for learning semantically rich text and image representations \cite{radford2021clip}.
The resulting text and image encoders perform well in many zero-shot tasks and have been successfully applied to image generation~\cite{ramesh2022hierarchical}, transfer learning~\cite{wortsman2022robust, chen2024understanding}, and multimodal large language models \cite{liu2023visual, beyer2024paligemma}.
CLIP-like models have been improved through refinements of encoder architectures, training data, and optimization schemes~\cite{cherti2023openclip, zhai2023sigmoid}.
However, these models are not designed to encode \textit{multimodal inputs} (an image-text pair) into a single feature vector, as text and images are processed by two separate encoders.
Several techniques have adapted pre-trained CLIP models for multimodal retrieval~\cite{liu2023univldr, wei2024uniir, zhang2024magiclens} or other downstream tasks~\cite{singh2022flava}. These methods typically merge features extracted by frozen unimodal encoders through either fixed functions or learnable modules. A different line of work trains multimodal sequence-to-sequence models \cite{wang2022ofa, lu2022unified, mizrahi20234m} using an encoder-decoder architecture~\cite{raffel2020exploring}. While well-suited for transfer learning, these models lack the strong zero-shot capabilities of \clip.

In this work, we propose a novel multimodal embedding method that extends \clip to multimodal inputs while preserving its strong vision-language alignment and zero-shot capabilities. %
Our first key novelty is to encode all input modalities (image, text, and their combinations) using a \textit{single~encoder}.
We achieve this by leveraging a discrete image tokenizer~\cite{yu2024titok}, which together with the standard text tokenizer maps inputs into unified sequences of tokens drawn from a \textit{finite multimodal vocabulary}.
Since the image tokenizer is trained solely for image compression and reconstruction, 
it does not introduce any bias with respect to text--image alignment.
Processing tokenized inputs with a single encoder, i.e., \textit{early~fusion}, allows modalities to interact at every encoding stage,
which differs from \textit{late fusion} where the deeper representations from unimodal encoders are merged \cite{liu2023univldr, zhang2024magiclens}.
Notably, we can train our architecture, named \ours, with a contrastive loss similar to standard \clip despite using a single encoder.
Moreover, discrete tokenizers enable us to seamlessly incorporate a masked multimodal modeling (MMM) loss into our training objective without the need for multiple auxiliary modules or the computational overhead incurred by prior work~\cite{singh2022flava}. %
Combining the MMM loss with the contrastive objective consistently enhances \ours{}’s performance across various zero-shot tasks (classification, retrieval, VQA, grounding), surpassing or matching late fusion baselines.

\looseness=-1
To comprehensively evaluate multimodal embedding models, we introduce novel tasks and datasets designed to test modality interactions.
We show that late fusion approaches struggle to solve tasks where visual information of the image is more relevant than semantic content, e.g. recognizing a correctly oriented crop.
Conversely, our early fusion architecture does not show this limitation, as both modalities communicate at every level of encoding after tokenization.
Finally, we demonstrate that training with \textit{hard negative} examples is essential for successfully learning these multimodal tasks.
Our work uncovers several interesting aspects of design, training, and evaluation of multimodal embedding models, and may impact future research in this area.

\textbf{Contributions.}
In summary, our work
\begin{itemize}[left=2mm, itemsep=0pt, parsep=3pt, topsep=0pt]
    \item introduces \ours, a novel multimodal embedding model, based on early fusion of discrete image and text tokens, processed by a single transformer encoder, which achieves performance surpassing or comparable to existing late fusion methods,
    \item shows that \ours can be effectively trained on both unimodal and multimodal data using a contrastive loss and incorporating hard negative examples, while natively supporting and significantly benefiting from a masked modeling objective,
    \item proposes novel evaluation tasks for multimodal embedding models, complementary to existing benchmarks, highlighting the importance of early fusion and hard negative examples. %
\end{itemize}

\section{Related Work}

\textbf{Unimodal embedding.}
Popular methods for vision-language pre-training such as \clip \cite{radford2021clip}, \siglip \cite{zhai2023sigmoid} and ALIGN \cite{jia2021scaling}  use separate networks to embed each modality.
An image encoder 
and a text encoder, 
with disjoint parameters, 
map data into a shared 
space.
Image-text pairs  with corresponding semantics are aligned via a contrastive 
loss on large image-caption datasets. These models achieve good zero-shot performance in tasks like image classification and retrieval, where inputs to be encoded
are from a single modality.

\textbf{Multimodal embedding.}
\looseness=-1
Certain tasks require encoding multimodal inputs into a single feature vector, which typical methods like \clip cannot directly handle. Alternative approaches merge representations from text and image encoders.
In \textit{score-level fusion}~\cite{liu2023univldr},  unimodal embeddings are simply summed (possibly with weighting).
This method avoids introducing additional merging modules. %
In contrast, \textit{feature-level fusion}~\cite{singh2022flava, zhang2024magiclens}
feeds unimodal embeddings to an additional network, 
typically a transformer architecture with multiple attention layers.
Moreover, BridgeTower~\cite{xu2023bridgetower} interconnects features within the latter blocks of the vision (\clip) and language encoders (RoBERTa): the resulting model is fine-tuned on various downstream tasks. 
\chst{\citet{caffagni2025recurrence} employ a recurrent cell that integrates textual and visual features from intermediate layers of a pre-trained CLIP model. 
Finally, VisualBERT~\cite{li2019visualbert} and UNITER~\cite{chen2020uniter} merge visual representations from pre-trained convolutional networks with text tokens and process these with a transformer, training for individual downstream tasks explicitly. In contrast, we use a discrete image tokenizer and focus on zero-shot evaluations.}

\textbf{Conversion of large multimodal models.}
Recent works \cite{jiang2024e5, jiang2025vlm2vec, gu2025breaking, xue2025improve}
convert autoregressive large multimodal models (LMMs) into encoders.
Such conversion is achieved by fine-tuning an LMM with contrastive learning to return semantically aligned feature vectors.
This strategy leverages the large pre-training datasets of the LMM to obtain rich representations, but comes with high inference costs due to their large size.%

\textbf{Multimodal embedding for composed image retrieval.}
Several embedding methods have been proposed for multimodal and composed image retrieval. One approach uses adapter modules \cite{saito2023pic2word, baldrati2023zero, gu2024lincir} or fine-tuning \cite{zhou2024vista} of a pre-trained vision encoder to map its output to a text encoder’s input space, enabling image-text encoding.
Notably, these approaches rely on pre-trained CLIP (and BERT) models, and are specialized for retrieval tasks.

\textbf{Early fusion in masked multimodal modeling.}
\citet{mizrahi20234m} and \citet{bachmann20244m21} train encoder-decoder architectures with masked modeling loss on multimodal data, using tokenized and pixel-based RGB images for vision datasets.
These models can generate data across modalities and adapt to different tasks via full fine-tuning, but lack zero-shot capabilities enabled by CLIP's contrastive pre-training.

\textbf{Early fusion in other domains.}
Early fusion has also been explored in other domains. Namely for image generation by DALL·E~\cite{ramesh2021dalle}, and for autoregressive models by Chameleon~\cite{team2024chameleon}. 
In contrast, our work focusses on embedding models and the direct comparison to late fusion baselines.

\begin{figure*}[t]
    \centering
    \def\imgheight{6.2cm}
    \begin{subfigure}{0.32\textwidth}
        \centering
        \includegraphics[height=\imgheight]{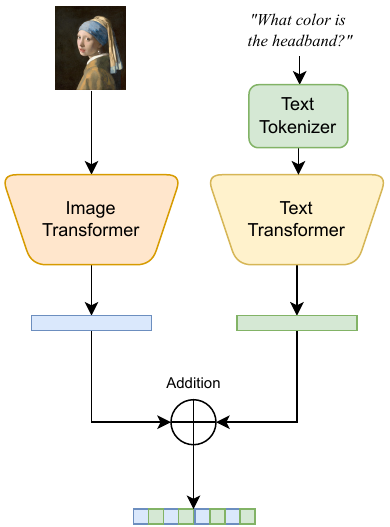}
        \caption{Score-level~\cite{liu2023univldr}}
        \label{fig:score-fusion}
    \end{subfigure}
    \hspace{2pt}
    \hfill
    \begin{subfigure}{0.32\textwidth}
        \centering
        \includegraphics[height=\imgheight]{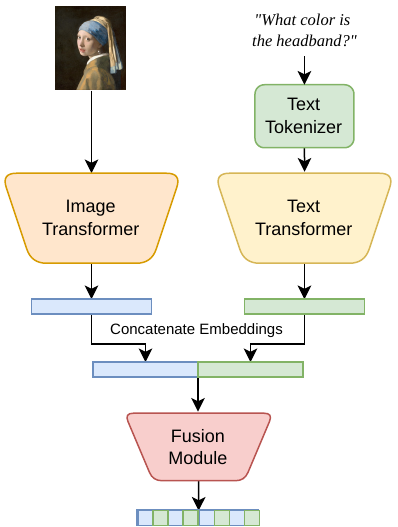}
        \caption{\magiclens~\cite{zhang2024magiclens}}
        \label{fig:magiclens}
    \end{subfigure}
    \hfill
    \begin{subfigure}{0.32\textwidth}
        \centering
        \includegraphics[height=\imgheight]{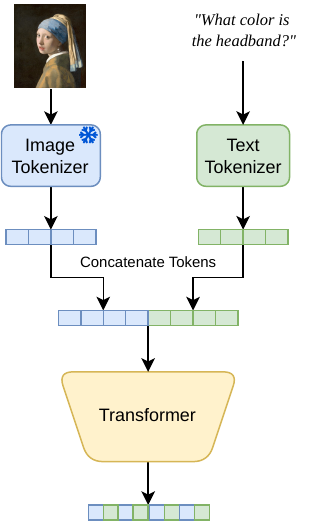}
        \caption{\ours (ours)}
        \label{fig:fuse-clip}
    \end{subfigure}

    \caption{\textbf{Comparison of architectures.} To obtain a multimodal embedding via contrastive learning, late fusion approaches first extract unimodal representations via unimodal encoders, then merge by addition \cite{liu2023univldr} or a fusion module \cite{zhang2024magiclens}. Conversely, our \ours uses a frozen image tokenizer to tokenize inputs of any modality into tokens from a unified vocabulary, which are then processed by a single encoder model. This approach leads to a simple architecture and early fusion of modalities.
    }
    \label{fig:architectures}
\end{figure*}

\section{\ours: Single-Encoder Multimodal Embedding via Early Fusion}

\subsection{Architecture}
\looseness=-1
Late fusion approaches merge image and text representations only in later layers, after %
independent encoding. Thus, each modality has limited influence on the final features since the modalities communicate only when already heavily processed.
However, a multimodal embedding should be strongly conditioned on both input modalities. For instance, in Fig.~\ref{fig:architectures}, the joint embedding of the image and the query \textit{``What color is the headband?''} should  align with that of \textit{``Blue''}, and exclude irrelevant details.
To achieve this, we propose \textit{(a)} enabling early interaction between modalities by merging them immediately after tokenization and \textit{(b)}~processing them with a single encoder.

\looseness=-1
\textbf{Tokenization.}
While text tokenization is straightforward, compressing images into tokens is more complex. We leverage recent progress in discrete image tokenizers \cite{vandenoord2017neural, esser2021taming}, and use those from the \titok family~\cite{yu2024titok} (frozen during our training), encoding each image into 128 discrete tokens. %
This ensures symmetry between image and text tokenization, allowing the transformer-based encoder to operate over a finite vocabulary.
This property allows us to use a masked multimodal modeling (MMM) loss,
without  ad-hoc tokenizers or extra computation, unlike \flava \cite{singh2022flava}. The MMM loss significantly improves model performance (see Sec.~\ref{sec:ablation}).
Moreover, \titok tokenizers are trained for image reconstruction on \imnet without text-guided semantic alignment and thereby avoid bias towards a specific representation that may arise during pre-training.  In contrast,  VISTA~\cite{zhou2024vista} initializes its vision encoder from CLIP, inheriting biases from contrastive pre-training.
Finally, \titok tokenizers provide high-quality compression, as shown  by their excellent reconstruction properties \cite{yu2024titok}.%

\textbf{Early fusion.}
Text and image are tokenized separately with a single vocabulary of non-overlapping text and vision tokens, obtained by extending the standard text vocabulary with the image token vocabulary.
The resulting token sequences are concatenated (images first, followed by text), with special beginning and end of text tokens (\texttt{<bot>}, \texttt{<eot>}) to separate modalities. %
Unimodal inputs omit the missing modality (an empty string is appended to image-only inputs).
The tokens are then mapped to $d$-dimensional vectors by an embedding matrix and combined with an additive positional embedding \revHYHX{that is applied jointly to the full concatenated sequence.
Finally, the sequence is processed by the unified transformer encoder.}

\textbf{Encoder.}
We adopt the transformer-based architecture of the \siglip text encoder, consisting of transformer blocks with bidirectional attention, \revXRX{thereby yielding a bidirectional encoder}. %
\revXRX{An attention mask is applied to exclude empty tokens from the self-attention computation.}
\revHYHX{We use the final-layer representation of the \texttt{<eot>} token as the output embedding, which aggregates information from all non-padding tokens through bidirectional attention.} %
Overall,  our model can be represented as
$f_{\vtheta_\textrm{tok}, \vtheta_\textrm{enc}}: I \times T \rightarrow \R^d$,
mapping a multimodal input $(\vi, \vt)$ (which may be unimodal when one modality is missing) to a $d$-dimensional feature vector. It is parameterized by the weights of the tokenizer $\vtheta_\textrm{tok}$ and encoder $\vtheta_\textrm{enc}$.

\textbf{Auxiliary prediction head.}
For the masked modeling loss, we introduce a classification head to predict masked tokens.
This module maps the output of the final transformer block to predictions over the token vocabulary $V$ and follows the FLAVA architecture~\cite{singh2022flava}: a two-layer network with shared embedding and unembedding matrices. We denote this head as $h_{\vtheta_\textrm{head}}:\R^d\rightarrow \R^{|V|}$, parameterized by $\vtheta_\textrm{head}$.

\subsection{Training objective}
\label{sec:training-fuselip}

\textbf{Contrastive loss.}
To match the zero-shot performance of popular language-image pre-training methods, we  optimize the sigmoid loss from \siglip~\cite{zhai2023sigmoid}. %
For a standard dual encoder with separate visual and text towers, and a batch $B$ of image-text pairs $\{(\vi_k, \vt_k)\}_{k=1}^{|B|}$, the \siglip loss is
\begin{align*}
    \hspace{-1mm}\L_\textrm{\siglip} =
    \frac{1}{|B|} \sum_{r=1}^{|B|}\sum_{s=1}^{|B|} \log\big(1 + e^{z_{rs} \left(-t \phi(\vi_r) \cdot \psi(\vt_s) + b\right)}\big),
\end{align*}
where  $\phi(\vi_r)$ and $\psi(\vt_s)$ are normalized embeddings from the vision and text encoders, $z_{rs}=1$  for positive pairs and~$-1$ otherwise, and $t,b$ are learnable parameters.
For \ours, the single multimodal encoder $f$ processes multimodal pairs $\{(\vz^1_k, \vz^2_k)\}_{k=1}^{|B|}$ into normalized embeddings, where $\vz^1_k$ and $\vz^2_k$ can be text, image, or image-text inputs.
The loss is then rewritten as
\begin{align*}
\hspace{-2mm}\L_\textrm{\siglip}^\textrm{MM} =
    \frac{1}{|B|} \sum_{r=1}^{|B|}\sum_{s=1}^{|B|} \log\big(1 + e^{z_{rs} \left(-t f(\vz^1_r) \cdot f(\vz^2_s) + b\right)}\big).
\end{align*}

\textbf{Masked modeling loss.}
Masked modeling is a popular pre-training approach for both text \cite{devlin2019bert} and image tasks \cite{bao2022beit}, where one aims to recover input tokens that have been masked.
FLAVA~\cite{singh2022flava} shows its effectiveness in multimodal embeddings.
However,  FLAVA relies on late fusion, using separate vision and text encoders merged by a multimodal module similar to a vision transformer \cite{dosovitskiy2021vit}.
Since the vision, text and multimodal embedding are provided by different branches of its architecture, a head for each input modality type is added to predict the masked tokens. %
Because its vision encoder produces continuous embeddings, an additional discrete tokenizer is required for masked modeling, increasing both computational cost and parameter count.

The \ours architecture simplifies the application of masked modeling loss %
to train multimodal encoders.
Since all input modalities are mapped to discrete tokens and processed by the same encoder, additional tokenizers and multiple prediction heads are not necessary.
Moreover, the masked modeling and contrastive losses are applied \textit{on the same masked input}, avoiding extra computational overhead.
As shown in Sec.~\ref{sec:ablation}, this strategy significantly improves performance across tasks. %
In practice, each token (except special tokens) is replaced with a \texttt{<MASK>} token with probability $p=0.1$ (\revXRX{this is distinct from the attention masking applied to empty input tokens}). Denoting $J(\vz)$ as the set of masked positions of the masked tokens for an input $\vz$, $Y(\vz)$ as the corresponding labels, and $f^{l}$ as the output of the last transformer block, the masked multimodal modeling (MMM) loss for the batch $B=\{(\vz^1_k, \vz^2_k)\}_{k=1}^{|B|}$ is
\begin{align*}
    \L_\textrm{MMM} = \frac{1}{|B|}\sum_{r=1}^{|B|} \sum_{i=1}^2 \sum_{(j, y) \in (J(\vz^i_r), Y(\vz^i_r))} \hspace{-14pt}
    \L_\textrm{CE}\left(h ( f^{l}_j(\vz^i_r)), y\right),
\end{align*}
with cross entropy loss $\L_\textrm{CE}$ and prediction head $h$.

\textbf{Final objective.}
The final optimization problem combines the two losses:
\( \min_{\vtheta_\textrm{enc}, \vtheta_\textrm{head}}\, \L_\textrm{\siglip}^\textrm{MM} + \alpha \,\L_\textrm{MMM}, \)
where $\alpha$ balances the two losses ($\alpha=0.25$ in all experiments).
The image tokenizer remains frozen during training, meaning that $\vtheta_\textrm{tok}$ is not optimized.
\revXRX{Both objectives are non-causal, therefore \ours remains a bidirectional encoder.}

\section{Training Data}
\label{sec:data}

\subsection{Unimodal data}
\label{sec:unimodal-data}
To train \ours and baseline models, we collect a variety of unimodal and multimodal data. %
We refer to image-text (I -- T) pairs data as unimodal since they do not require
joint encoding of inputs from different modalities. These datasets are commonly used to train CLIP-like models, and can also be leveraged by multimodal encoders.
We use \cciii \cite{sharma2018conceptual} and \ccxii \cite{changpinyo2021cc12m}, as they provide high-quality images and captions, and are amenable to training within academic compute constraints.

\begin{figure*}
\centering
\footnotesize
\tabcolsep=1pt
\def\imgwidth{0.08\linewidth}
\begin{tabular}{m{4mm}c@{\hspace{2mm}}c@{\hspace{0.5mm}}@{\hspace{0.5mm}}*{9}{c}}
\toprule
& \small Query & \multicolumn{4}{c}{\small Retrieval pool} && \small Query  & \multicolumn{4}{c}{\small Retrieval pool} \\
\midrule
 & \multicolumn{3}{l}{The dog on the left} &&&& \multicolumn{3}{l}{The toy on the right} \\
\multirow[t]{1}{*}[1mm]{\rotatebox{90}{\textbf{\oipos}}} & \includegraphics[width=\imgwidth]{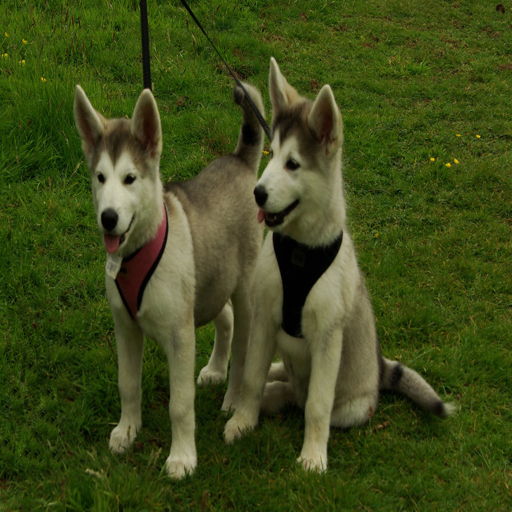} &
\includegraphics[width=\imgwidth,cframe=red 1pt]{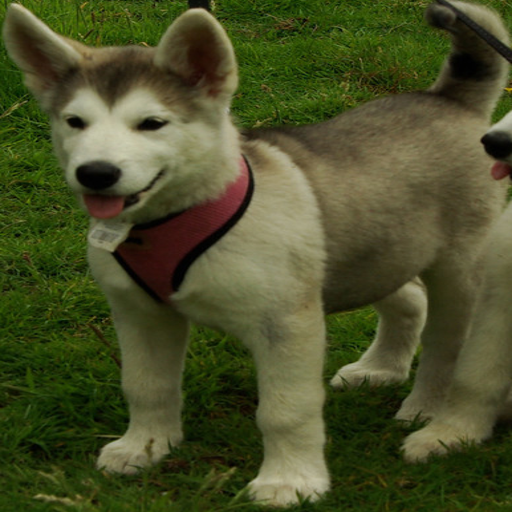} &
\includegraphics[width=\imgwidth]{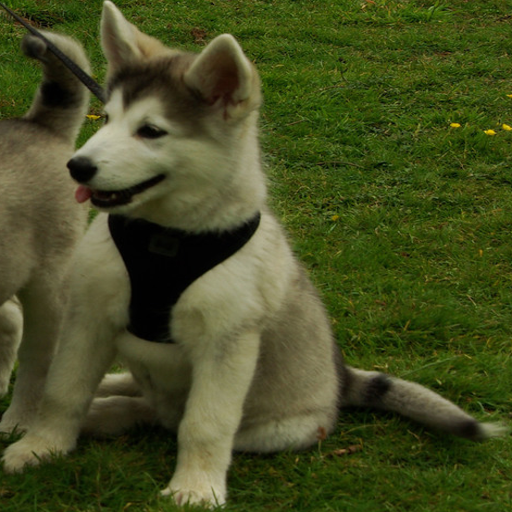} & \includegraphics[width=\imgwidth]{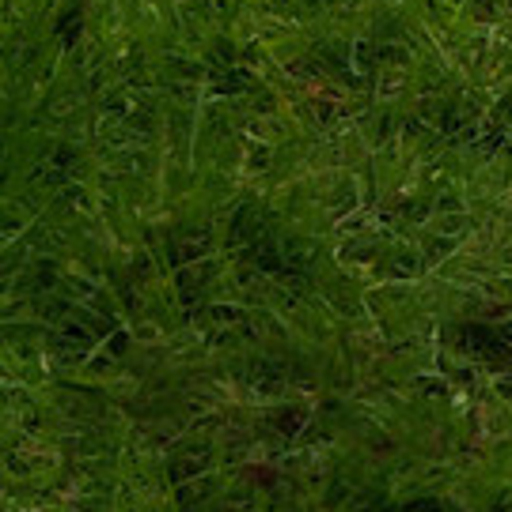} & \includegraphics[width=\imgwidth]{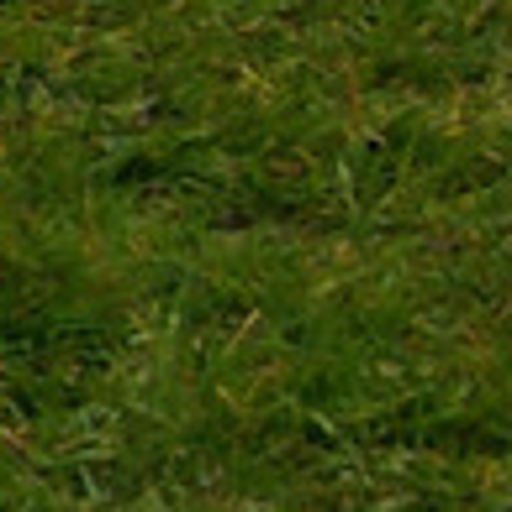} & &
\includegraphics[width=\imgwidth]{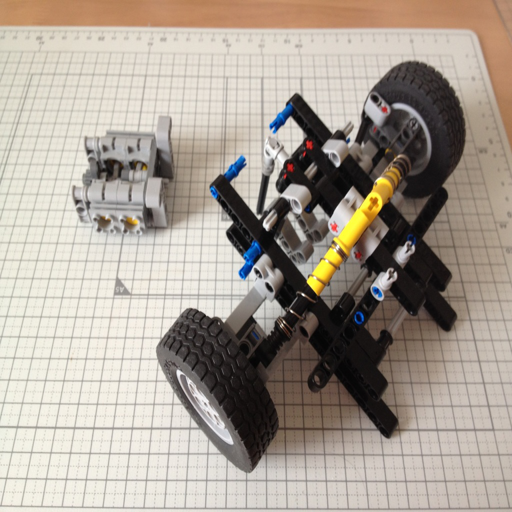} & \includegraphics[width=\imgwidth,cframe=red 1pt]{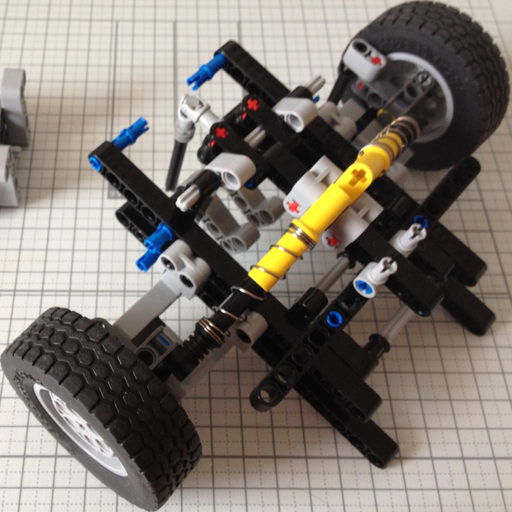} & \includegraphics[width=\imgwidth]{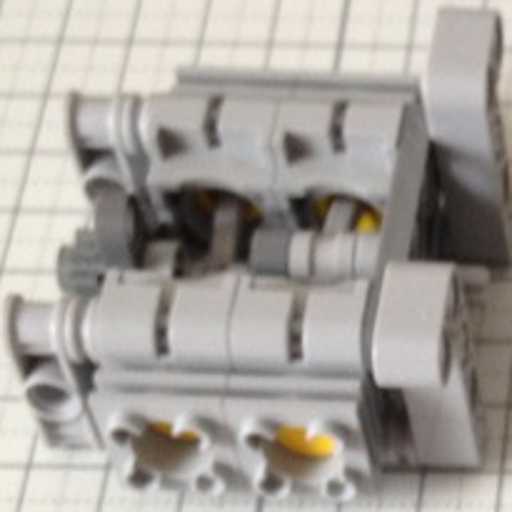} & \includegraphics[width=\imgwidth]{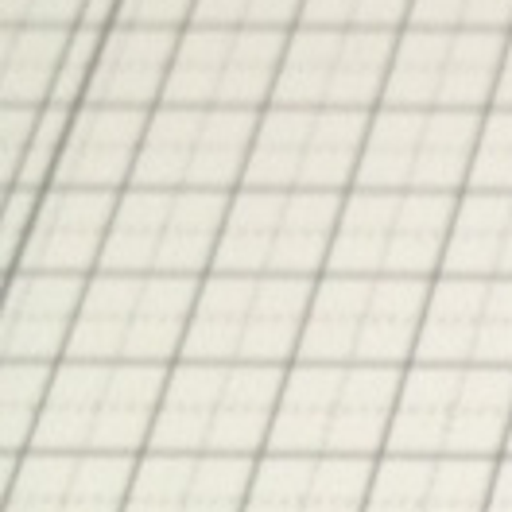} & \includegraphics[width=\imgwidth]{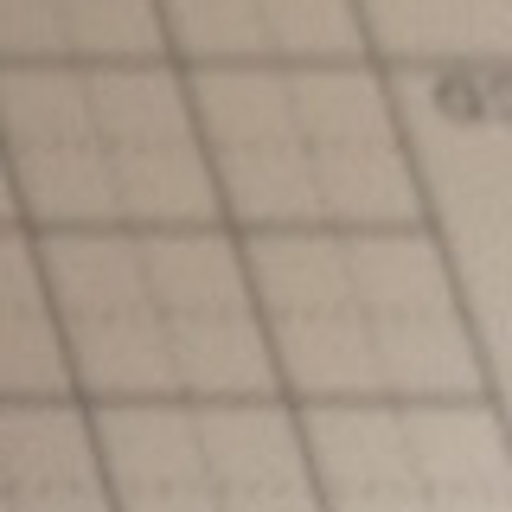} \\
\midrule
\midrule
& \small Query & \multicolumn{9}{c}{\small Retrieval pool} \\
\midrule
& Taxi \\
\multirow[t]{1}{*}{\rotatebox{90}{\textbf{\oicrop}}} & \includegraphics[width=\imgwidth]{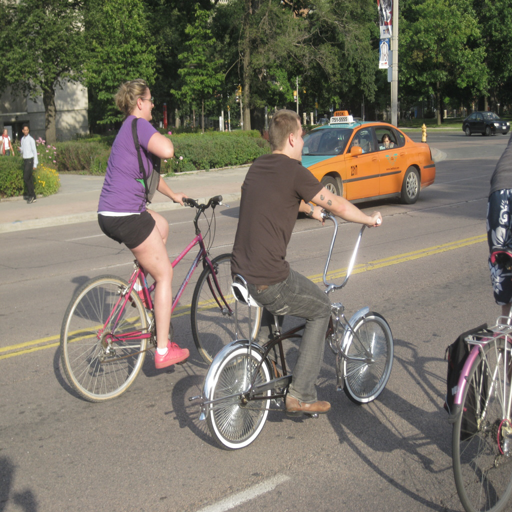} & \includegraphics[width=\imgwidth,cframe=red 1pt]{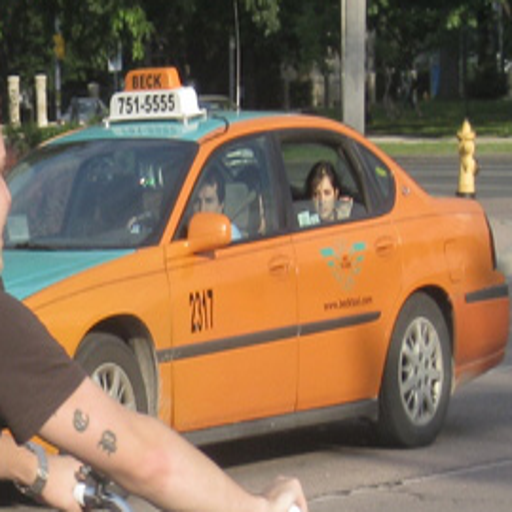} & \includegraphics[width=\imgwidth]{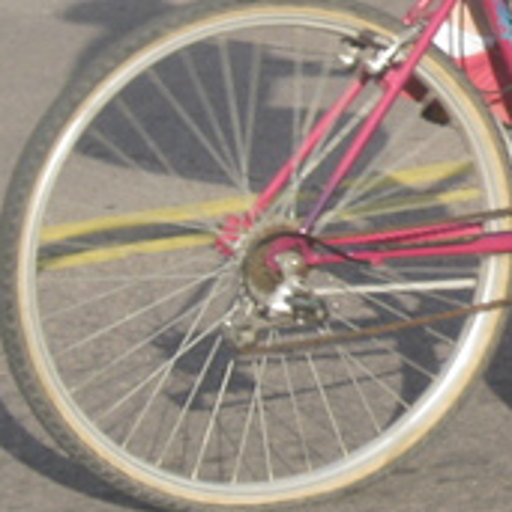} & \includegraphics[width=\imgwidth]{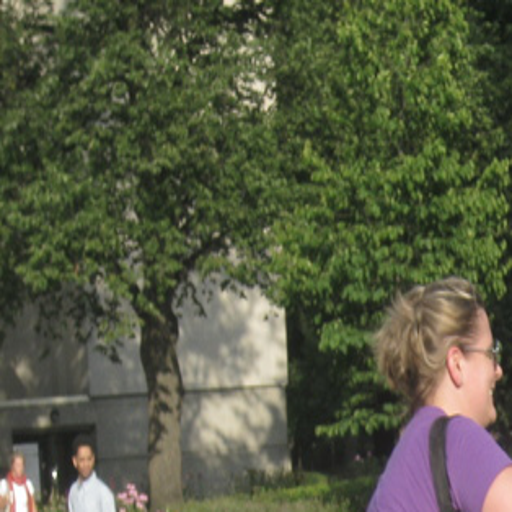} & \includegraphics[width=\imgwidth]{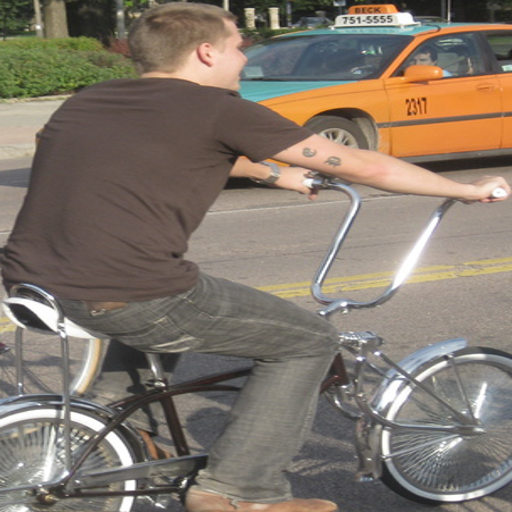} & \includegraphics[width=\imgwidth]{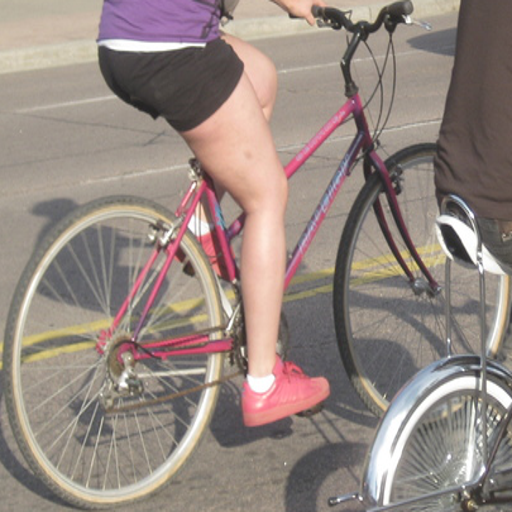} & \includegraphics[width=\imgwidth]{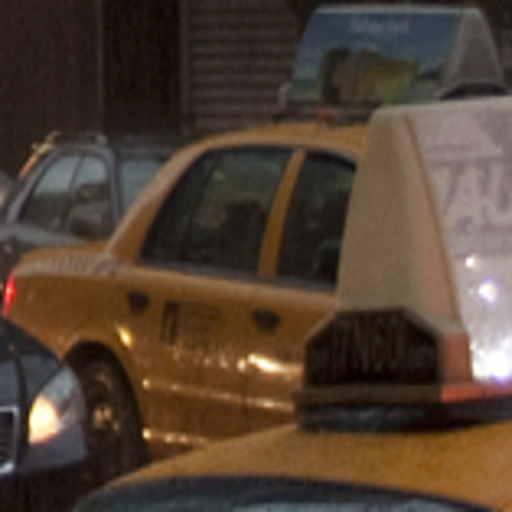} & \includegraphics[width=\imgwidth]{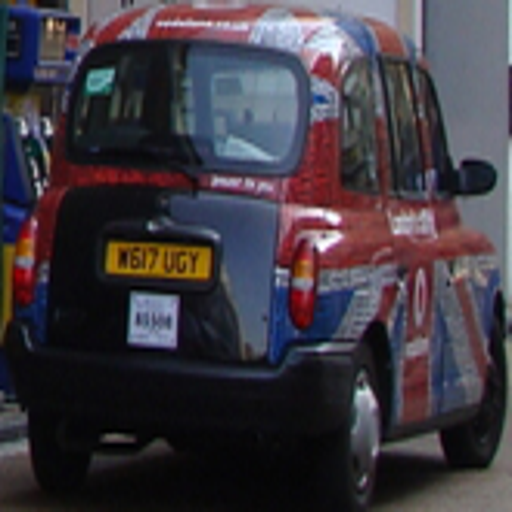} & \includegraphics[width=\imgwidth]{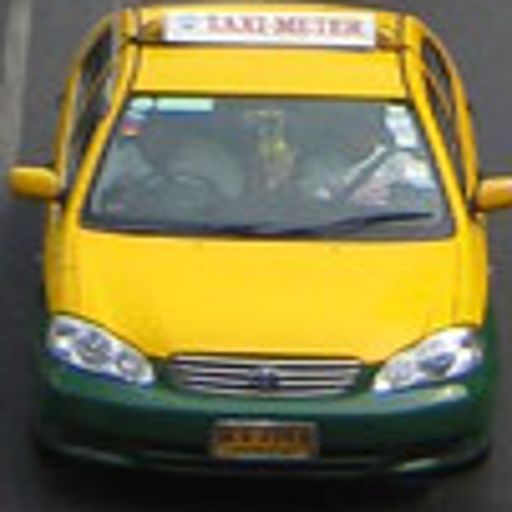} & \includegraphics[width=\imgwidth]{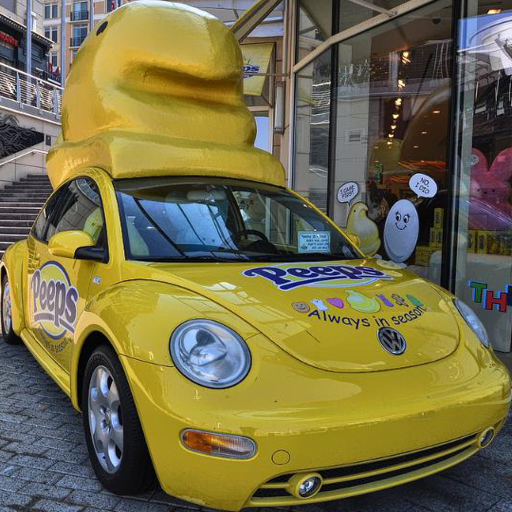} & \includegraphics[width=\imgwidth]{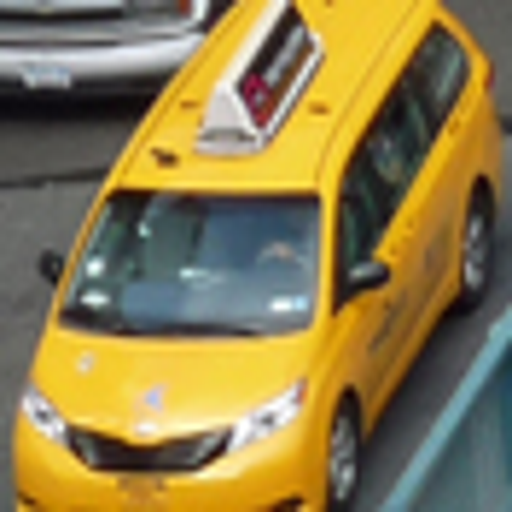} \\
\midrule
\midrule
& \multicolumn{5}{l}{headlight on a motorcycle } \\
\multirow[t]{1}{*}{\rotatebox{90}{\textbf{\vgcrop}}} & \includegraphics[width=\imgwidth]{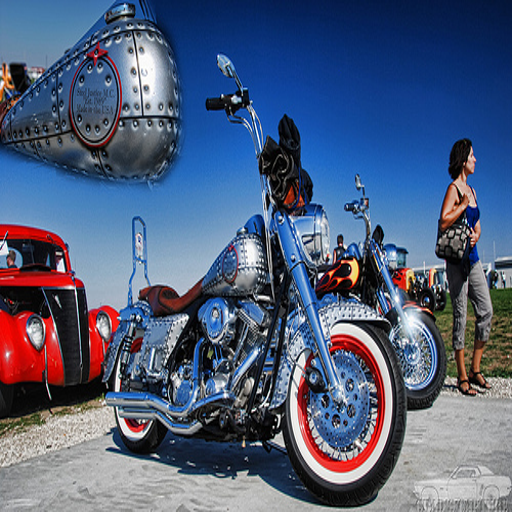} & \includegraphics[width=\imgwidth,cframe=red 1pt]{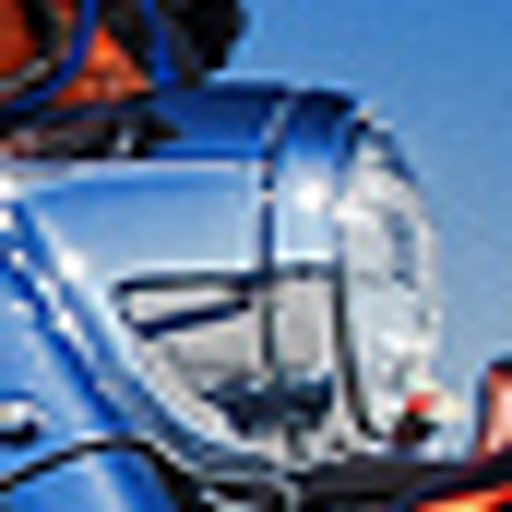} & \includegraphics[width=\imgwidth]{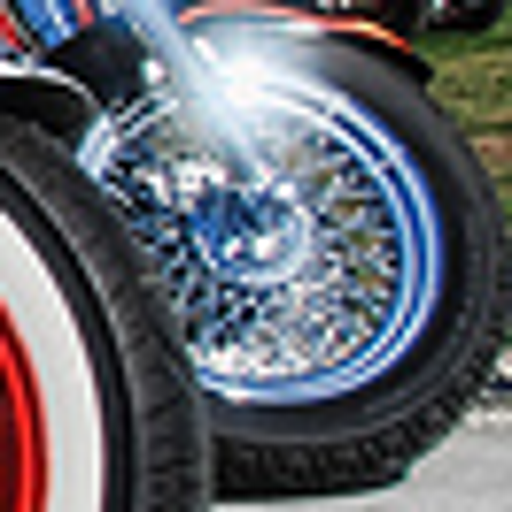} & \includegraphics[width=\imgwidth]{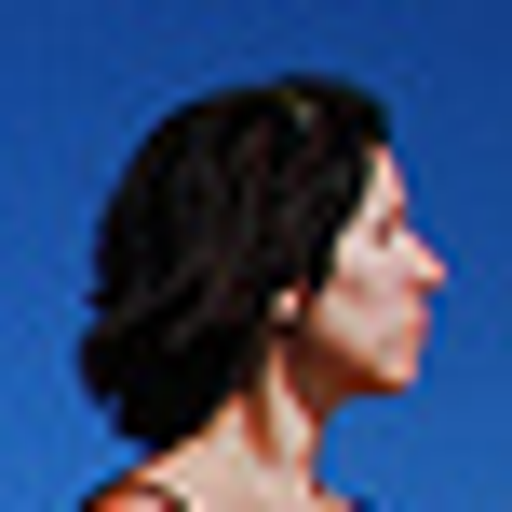} & \includegraphics[width=\imgwidth]{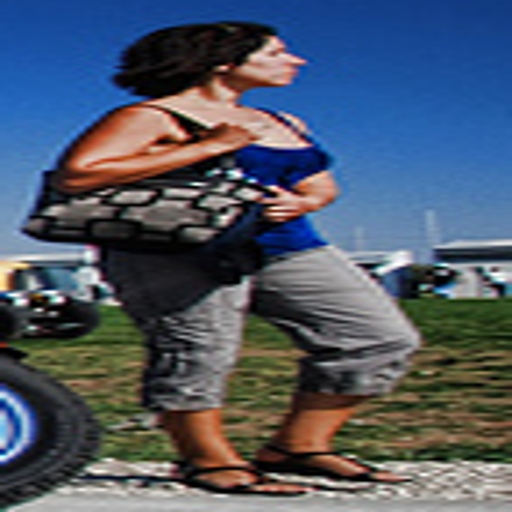} & \includegraphics[width=\imgwidth]{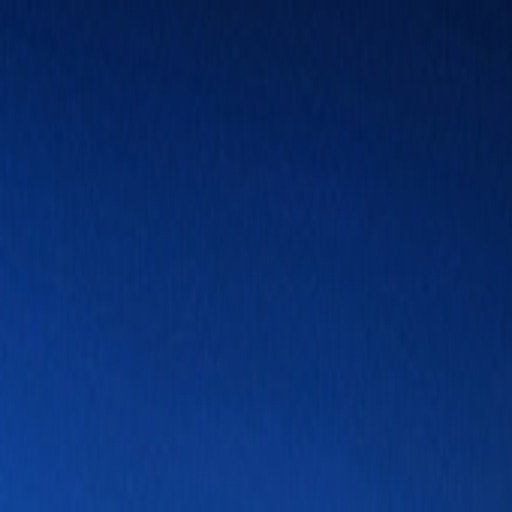} & \includegraphics[width=\imgwidth]{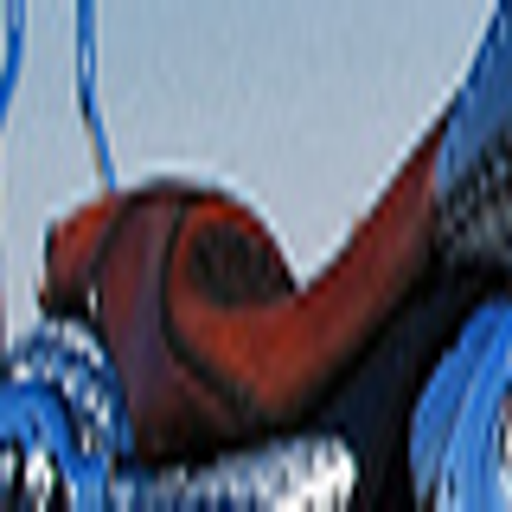} & \includegraphics[width=\imgwidth]{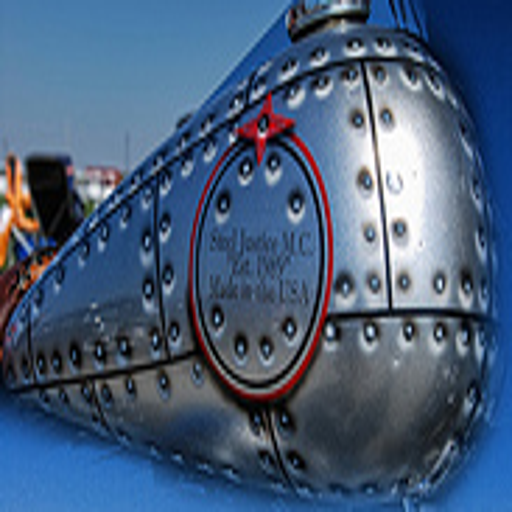} & \includegraphics[width=\imgwidth]{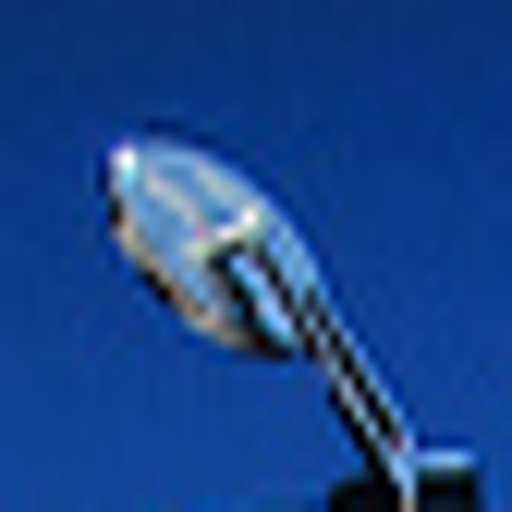} & \includegraphics[width=\imgwidth]{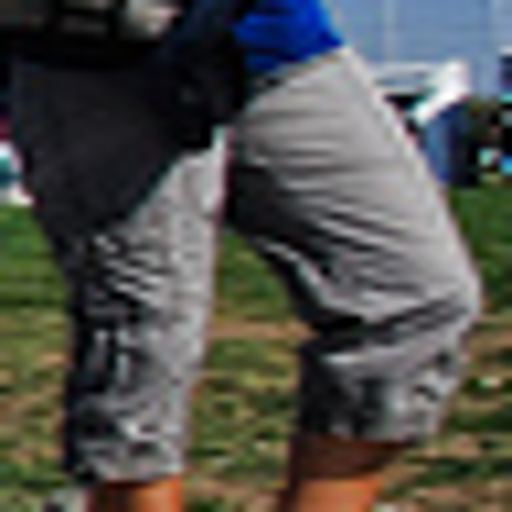} & \includegraphics[width=\imgwidth]{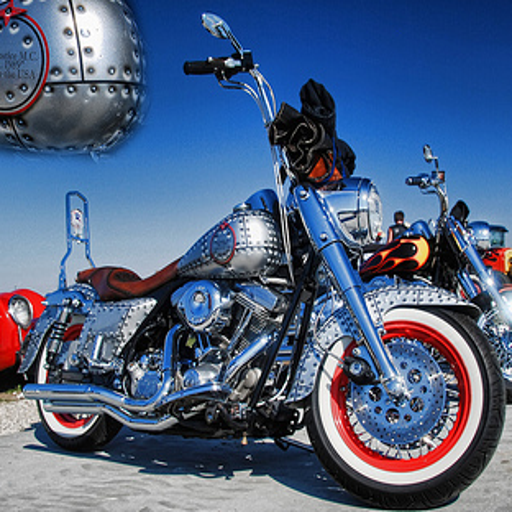} \\
\bottomrule
\end{tabular}
\caption{\textbf{\oipos, \oicrop and \vgcrop tasks.} We show examples of these tasks (described in Sec.~\ref{sec:eval-tasks}). 
The retrieval pool of \oicrop comprises crops from the same image, as well as the target object cropped from other images. For \oipos and \vgcrop it contains only crops of the query image. 
We show the whole retrieval pool for \oipos, \oicrop, and a sample for \vgcrop. Red frames highlight the ground-truths.
}
\label{fig:oi-vg-crop-tasks}
\end{figure*}

\subsection{Multimodal data}
\label{sec:multimodal-data}
Datasets with multimodal inputs are scarce.
Thus, besides relying on existing ones, we generate additional multimodal data from unimodal data, effectively scaling dataset size at minimal 
additional collection cost.

\textbf{Text-guided image transformations (TGIT).}
We generate multimodal data from image datasets by applying transformations and describing them in text.
Specifically, we form pairs $(\vi\oplus \vt, \vi^\prime)$, where $\vi$ is the original image, $\vt$ is the %
transformation description in natural language, and $\vi^\prime$ is the transformed image. We consider %
transformations such as
\textit{random cropping}, \textit{random rotations}, \textit{flipping}, \textit{colorization}, and \textit{color jittering}.
See App.~\ref{app:data} for a detailed description and examples.
We apply this approach to a subset of \cciii and \ccxii but it can be extended to any image dataset.
Since text prompts describe only the transformations, without semantic information about the image, the model must rely on both modalities  and cannot solve the task using only text or image.

\textbf{VQA data generated from image-text datasets (\cciiivqa).}
We generate VQA data from \cciii by prompting an LLM (Llama-3.1-8B-Instruct from~\citet{dubey2024llama}) to rewrite captions as question-answer pairs using a structured system prompt (see Fig.~\ref{fig:system_prompt} in Appendix).
We generate 2.4M VQA samples. Notably, this method is scalable to any image-caption dataset.
While generating VQA from image-text pairs was explored by \citet{changpinyo2022all}, we leverage the recent advancements in LLMs for a simplified and unsupervised generation pipeline.

\textbf{VQA data from \vig (\vgvqa).}
We use the existing VQA samples from \vig (VG)~\cite{krishna2017visualgenome}, denoted as \vgvqa, in the standard IT -- T format.

\textbf{Visual Grounding with  \vig (\vgcrop).}
VG contains images with rich annotations, in particular natural language descriptions of image regions that are bounded by rectangular boxes. We use these descriptions to build a training dataset, referred to as \vgcrop. Namely, given an image $\vi$ and a textual description $\vd$ of some region $\vi^\prime$ in $\vi$, we form pairs $(\vi \oplus \vd, \vi^\prime)$. That is, the model is tasked to find the crop of the image, given a natural language instruction. Thus, the modality combination is \mbox{IT -- I}, %
which is complementary to that of VQA datasets.

\textbf{HQ-Edit.}
Finally, \hqedit~\cite{hui2024hqedit} consists of synthetically generated image edits. We integrate this dataset into our training by tasking the model to find the correctly edited image. Given an image $\vi$, an edit described in natural language $\ve$, and the edited image $\vi^\prime$, we form pairs $(\vi \oplus \ve, \vi^\prime)$. Using the inverse edits $\ve^\prime$ that are contained in \hqedit, we also form the corresponding inverse pairs $(\vi^\prime \oplus \ve^\prime, \vi)$.
As \hqedit also contains captions $\vc^\prime$ of the edited images, we build additional training samples as $(\vi \oplus \ve, \vi^\prime \oplus \vc^\prime)$. This covers both IT -- I and IT -- IT modalities.

\subsection{Training with hard negatives}
\label{sec:training-scheme}

For training, we merge all datasets described in Sections~\ref{sec:unimodal-data} and~\ref{sec:multimodal-data}, ensuring that batches contain diverse tasks and modalities.
Due to the contrastive nature of the \siglip loss, each query is contrasted against all non-matching targets in the batch.
\revHYHX{
Thus, the choice of batch composition determines which negatives the model observes during training. 
}%
Closely related samples, called \textit{hard negatives}, have been shown to improve contrastive learning~\cite{kalantidis2020hard,robinson2020hard,zhang2024magiclens}.
We design and integrate hard negatives into the training of \ours and baselines by ensuring that batches contain semantically similar examples.
\revHYHX{
As a result, non-matching cross-combinations between related samples act as hard negatives: the model must, for example, distinguish the correct transformation among several transformations of the same image, or the correct region among several regions from the same image.
}%
Specifically, for \cciiiaug and \ccxiiaug we sample multiple transformations of the same image, for \vgcrop and \vgvqa each batch includes three additional samples from the same query image, with different descriptions or questions, and for \hqedit we include for every sample the corresponding inverse edit sample in the batch.
We  show in Sec.~\ref{sec:ablation} the key role of hard negatives for learning these tasks.

\begin{table}[t]
\centering
\begin{minipage}[t]{0.48\columnwidth}
\centering
\small
\setlength{\tabcolsep}{3pt}
\caption{\textbf{Parameter comparison} (in million).\\
Greyed numbers indicate non-trainable parameters.}
\vspace{2pt}
\label{tab:params}
\begin{tabular}{l  ccc  cc}%
    \toprule
     Model & Image & Text & Fusion & \cellcolor{myBlue} Trained & Total \\
     \midrule
     \siglipSsf & 21.8 & 40.5 & - & \cellcolor{myBlue} 62.3 & 62.3 \\
      \siglipSmlf   & 21.8 & 40.5 & 7.7 & \cellcolor{myBlue} 70.0 & 70.0 \\
      \siglipBsf & 86.2 & 63.5 & - & \cellcolor{myBlue} 149.7 & 149.7 \\
      \siglipBmlf & 86.2 & 63.5 & 13.7 & \cellcolor{myBlue} 163.4 & 163.4 \\
      \midrule
     \oursS & \textcolor{gray!70}{25.9} & - & 42.1 & \cellcolor{myBlue} 42.1 & 68.0 \\
     \oursB & \textcolor{gray!70}{86.6} & - & 65.6 & \cellcolor{myBlue} 65.6 & 152.2 \\
     \bottomrule
\end{tabular}
\end{minipage}
\hfill
\begin{minipage}[t]{0.48\columnwidth}
\centering
\small
\setlength{\tabcolsep}{2pt}
\caption{\textbf{Number of samples per dataset.}
We show the number of samples (in million) of the training datasets
described in Sec.~\ref{sec:data}, and the combination of modalities
that each dataset exhibits.}
\label{tab:datasets}
\begin{tabular}{@{}lc@{\hspace{2mm}}cccccccc@{}}
    \toprule
     \multirow{2}{*}{Setting} & CC & TGIT & \makecell{CC3M-\\VQA} & \makecell{VG-\\VQA} & \makecell{VG-\\Crop} & \makecell{HQ-\\Edit} \\
     \cmidrule{2-7}
     & I -- T & IT -- I & IT -- T & IT -- T & IT -- I & IT -- I(T) \\
     \midrule
     \cciii & 2.6 & 0.3 & 2.4 & 0.7 & 5.4 & 0.3 \\
     \ccxii & 10.6 & 0.3 & 2.4 & 1.4 & 5.4 & 0.3 \\
     \bottomrule
\end{tabular}
\end{minipage}
\end{table}

\section{Experiments}
\label{sec:experiments}

\textbf{Models.} 
We train two versions of \ours: \oursS uses the \titok-S tokenizer and a small transformer %
as implemented by OpenCLIP \cite{cherti2023openclip}, while \oursB uses the \titok-B tokenizer and a base transformer.
We consider two late fusion baselines: \textit{(1)} score fusion (SF), where the multimodal embedding is obtained by summing the unimodal embeddings from text and vision encoders, and \textit{(2)} \magiclens fusion (MLF)~\cite{zhang2024magiclens}, which uses a transformer module to merge the unimodal embedding vectors (see Fig.~\ref{fig:architectures}). %
Both baselines use text and vision encoders from \clip with ViT-S or ViT-B architecture~\cite{dosovitskiy2021vit}, 
and are trained on the \siglip loss using the same datasets and hard-negatives as \ours. 
We denote them as \siglipsf and \siglipmlf respectively. As we train all models from scratch, we do not compare against methods that fine-tune pre-trained models \cite{zhang2024magiclens,jiang2025vlm2vec}.
In Table~\ref{tab:params}, we report the parameter count for all architectures: \oursS has a total number of parameters similar to the S-sized baselines, but significantly fewer trainable ones as the image tokenizer remains frozen. In contrast, the total parameter amount of \oursB roughly matches that of B-sized baselines, while the trainable parameter amount of \oursB is similar to the S-sized baselines.
This discrepancy in trainable versus total parameters also affects the training cost: 
training of \ours is faster and requires less GPU memory as shown in App.~\ref{app:model-details}.

\textbf{Training.}
We train 8 epochs (total of 93M seen samples) on \cciii plus multimodal data (\cciiimm), and on \ccxii plus multimodal data (\ccxiimm) we train 16 epochs (326M samples), see Table~\ref{tab:datasets}.
Full training hyperparameters are provided in App.~\ref{app:training}.

\subsection{Multimodal evaluation tasks}
\label{sec:eval-tasks}

For evaluation, %
we consider a variety of
embedding tasks. Besides testing on \vgcrop and \cciiiaug, used during training, we collect existing and new datasets 
described here (details in App.~\ref{app:eval-tasks}).
The diversity of these tasks enables a more comprehensive evaluation of different model capabilities.
\revHYHX{%
On each task, given a query, the model has to retrieve the correct target from a set of candidate targets, as detailed below.
Queries and candidates are embedded and the model selects the the candidate with the highest cosine similarity to the query in the embedding space.
}%

\textbf{Massive Multimodal Embedding Benchmark.}
MMEB~\cite{jiang2025vlm2vec} is a multimodal embedding benchmark consisting of 36 subtasks that are split into the categories \textit{Classification}, \textit{VQA}, \textit{Retrieval}, and \textit{Grounding} and cover multiple modalities. Many of the subtasks are considered standard tasks in their domain. Each subtask includes 1000 samples: for each sample the model has to select the correct answer among 1000 candidates, 
\revHYHX{%
except some classification tasks (e.g.~CIFAR-10) contain fewer candidates.
We report the mean top-1 accuracy over all subtasks within each category.
}%

\textbf{Visual Grounding with OpenImages.}
We leverage OpenImages~\cite{OpenImages} to create a new task where the goal is to select the correct crop of an image given a query text.
For \textbf{\oicrop} the query text is the name of an object 
in the image. As negatives, we include four crops of other objects from the query image and five crops of the same object from other images (in contrast to \vgcrop which uses only crops of the same image). %
For \textbf{\oipos}, the model has to select the correct crop of an object from an image that contains this object exactly twice, given an instruction to select the left/right instance. The retrieval pool includes crops from the outer parts of the query image as decoy.
We give a detailed description of the data collection method in App.~\ref{app:eval-tasks}, which results in 1046 and 2546 samples respectively (see Fig.~\ref{fig:oi-vg-crop-tasks}).

\textbf{\vgcrop.}
We %
use 1574 validation samples from \vgcrop (Sec.~\ref{sec:multimodal-data}), using all existing regions in each sample as the respective candidates (see Fig.~\ref{fig:oi-vg-crop-tasks}). We prune regions that significantly overlap (IoU over 0.3) to facilitate unambiguous tasks. The size of the retrieval pool depends on the amount of available region descriptions, with an average of 15.9 images. %

\textbf{\cciiiaug.}
We use the validation split of \cciiiaug to test models on retrieving the correctly transformed image. For \textit{crop} and \textit{rotation} the retrieval pools consists of all possible transformations (9 and 18), for \textit{jitter} we use 10, for \textit{flip} the original and horizontally/vertically flipped images (3), and for \textit{colorize} the original and the target sample (2). Each subtask is evaluated on 1000 samples.

\textbf{\imnet.}
\revHYHX{%
We evaluate zero-shot classification on the full \imnet-1k~\cite{deng2009imagenet} validation set. Following \clip~\cite{radford2021clip}, we use the ensemble of 80 prompt templates per class, average the corresponding text embeddings, and assign each image to the class with highest cosine similarity. This differs from the \imnet evaluation included in MMEB, which uses only 1k validation samples and a single text prompt template per class (``a photo of a {class}'').
}%

\begin{table*}[t]
\small
\centering
\tabcolsep=2.pt
\caption{%
\looseness=-1
\textbf{Evaluation on embedding tasks.}
We report accuracy of late fusion baselines (\siglipsf, \siglipmlf)  and \ours models, trained on \cciii or \ccxii plus multimodal data.
Our \oursB achieves the best results across nearly all tasks, despite having fewer trainable parameters than \siglip-B. We observe large margins on \oipos and TGIT, even for \oursS, indicating that early fusion better captures image-text relations. \emph{\zs{}zero-shot}
}
\label{tab:mm-embedding}
\begin{tabular}{l lccccccccc
}
    \toprule
    Training & Model & Classification\zs & VQA\zs & Retrieval\zs & Grounding\zs & ImageNet\zs & VG-Crop & OI-Crop\zs & OI-Pos\zs & TGIT \\
    \midrule
     &\siglipSsf & 21.5 & 12.7 & 13.0 & 74.8 & 8.8 & 52.0 & 55.2 & 45.4 & 57.3 \\
    &%
    \siglipSmlf & 18.0 & 14.2 & 12.7 & 74.2 & 10.2 & 53.0 & 66.2 & 46.9 & 67.2 \\
    & \siglipBsf & 22.2 & 13.6 & 13.4 & 77.2 & 10.3 & 55.1 & 56.9 & 45.9 & 56.6 \\
    & \siglipBmlf  & 19.5 & 14.8 & 13.9 & 76.9 & 12.2 & 55.4 & \textbf{68.4} & 47.4 & 69.4 \\
    \cmidrule(l){3-11}
    \rowcolor{myBlue} \cellcolor{white} & \oursS & 18.5 & 15.9 & 11.2 & 70.8 & 13.5 & 49.6 & 59.8 & 53.9 & 79.0 \\
    \rowcolor{myBlue} \cellcolor{white} \multirow{-7}{*}{\makecell{\cciii\\+\mmdata}} &\oursB & \textbf{23.3} & \textbf{17.5} & \textbf{15.0} & \textbf{82.4} & \textbf{18.1} & \textbf{55.8} & 68.1 & \textbf{70.8} & \textbf{94.3} \\
    \midrule
    & \siglipSsf & 30.4 & 16.2 & 23.8 & 74.2 & 21.4 & 57.1 & 60.1 & 47.1 & 66.0 \\
    &\siglipSmlf & 28.5 & 16.9 & 23.2 & 72.7 & 25.5 & 58.8 & 72.2 & 46.6 & 81.0 \\
    & \siglipBsf & \textbf{31.5} & 17.0 & 23.8 & 72.7 & 25.4 & 58.0 & 63.2 & 47.3 & 67.1\\
    & \siglipBmlf & 30.3 & 16.8 & 23.2 & 73.4 & 28.8 & \textbf{61.5} & \textbf{74.0} & 48.9 & 78.1 \\
    \cmidrule(l){3-11}
    \rowcolor{myBlue} \cellcolor{white} &\oursS & 25.2 & 18.2 & 20.1 & 75.2 & 26.0 & 53.5 & 64.7 & 61.5 & 90.6 \\
    \rowcolor{myBlue} \cellcolor{white} \multirow{-7}{*}{\makecell{\ccxii\\+\mmdata}} &\oursB & 31.2 & \textbf{19.8} & \textbf{26.2} & \textbf{82.3} & \textbf{32.7} & \textbf{61.5} & 71.3 & \textbf{68.9} & \textbf{94.2} \\
    \bottomrule
\end{tabular}
\end{table*}

\subsection{Main results}

Table~\ref{tab:mm-embedding} reports the performance of the various models on the evaluation tasks detailed above. %
First, \oursB achieves the best results across nearly all tasks and training data configurations,
often with large margin.
It attains highest scores on 8 respectively 7 out of 9 benchmarks in the \cciiimm and \ccxiimm training configurations.
While the total parameters of \oursB are similar to B-sized baselines, it has significantly fewer trainable ones (Table~\ref{tab:params}).
Notably, the non-trainable parameters come from the frozen image tokenizer, which is trained for image reconstruction and does not contribute directly to image-text alignment.
\revMAoJ{
We emphasize that this comparison concerns trainable parameters and hence optimization and training cost. At inference time, the frozen tokenizer is still part of \ours and must be evaluated for image-containing inputs (see Table~\ref{tab:runtime} for a runtime comparison).
}%

\revHYHX{
Second, FuseLIP-S shows a more mixed picture. While it has a similar total parameter count to the S-sized baselines, it has substantially fewer trainable parameters because the image tokenizer is frozen~(see Table 1). 
This reduced trainable capacity, combined with the lower fidelity image tokenizer, likely makes FuseLIP-S less competitive on some standard embedding tasks, such as Classification and Retrieval. 
}%
\revXRX{
Nevertheless, FuseLIP-S already improves over the S- and B-sized late-fusion baselines on several tasks that require joint image-text reasoning, in particular VQA, OI-Pos, and TGIT. Scaling to FuseLIP-B, which uses both a stronger tokenizer and a larger transformer encoder, yields the strongest overall results.
}%
As \ours significantly outperforms late fusion models, especially score fusion, on \cciiiaug,
we analyze this key result in more detail below.
Overall, these results suggest that early fusion of discrete tokens, even with a single encoder, is highly effective  for aligning multimodal representations.

\begin{figure*}[!t]
\centering
\footnotesize
\def\imgwidth{0.133\linewidth}
\tabcolsep=1pt
\small
\begin{tabular}{c@{\hspace{5pt}}cccccc}
\toprule
\small Query & \small \siglipSsf & \small \siglipSmlf & \small \siglipBsf & \small \siglipBmlf & \small \oursS & \small \oursB \\
\midrule
Horizontal flip & \xmark & \xmark & \xmark & \cmark & \cmark & \cmark \\
\includegraphics[width=\imgwidth]{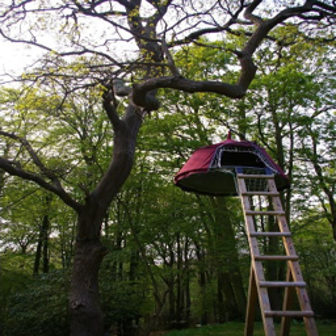} & \includegraphics[width=\imgwidth]{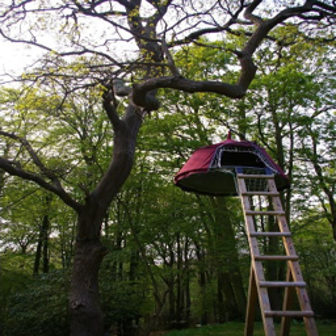} & \includegraphics[width=\imgwidth]{assets/aug-fig-neurips/flip/sample0000_ret001.png} & \includegraphics[width=\imgwidth]{assets/aug-fig-neurips/flip/sample0000_ret001.png} & \includegraphics[width=\imgwidth]{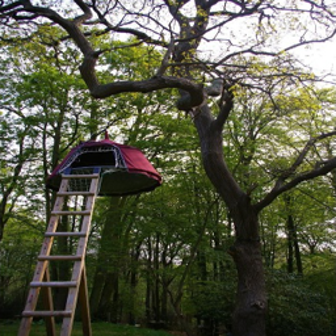} & \includegraphics[width=\imgwidth]{assets/aug-fig-neurips/flip/sample0000_ret000.png} & \includegraphics[width=\imgwidth]{assets/aug-fig-neurips/flip/sample0000_ret000.png} \\
\bottomrule
\end{tabular}
\caption{\revMAoJ{\textbf{TGIT evaluation examples.} We illustrate the ``horizontal flip'' task on models trained on \ccxiimm. While several baselines select the original image, only \mbox{\siglipBmlf} and both \ours variants return the correctly flipped image. More examples are shown in Fig.~\ref{fig:cc3m-aug-eval}.
}}
\label{fig:cc3m-aug-eval-short}
\end{figure*}
\begin{wraptable}{r}{0.48\columnwidth}
\small
\centering
\caption{\textbf{Breakdown over TGIT tasks.}
We report accuracy (\%) on five text-guided transformations for models trained on \ccxiimm.
Only \ours{} variants solve \textit{Crop}, \textit{Rotate} and \textit{Flip}.}
\label{tab:cc3m-aug}
\vspace{1mm}
\tabcolsep=4.5pt
\begin{tabular}{l cccccccc
}
    \toprule
    Model & Crop & Rotate & Flip & Jitter & Color %
    \\
    \midrule
    \siglipSsf & 49.7 & 84.7 & 15.6 & 80.7 & 99.4 \\
    \siglipSmlf & 75.8 & 83.3 & 56.4 & 89.7 & 100.0 \\
    \siglipBsf & 48.5 & 84.6 & 18.5 & 84.2 & 99.7 \\
    \siglipBmlf & 55.0 & 85.8 & 58.8 & 90.8 & 99.9 \\
    \rowcolor{myBlue} \oursS & 97.4 & 81.8 & 92.1 & 82.2 & 99.6 \\
    \rowcolor{myBlue} \oursB & 99.7& 94.7 & 89.2 & 87.7 & 99.6 \\
    \bottomrule
\end{tabular}
\end{wraptable}

\textbf{Why early fusion helps to understand text-guided transformations.}
Our models achieve the largest improvements over baselines on \cciiiaug, where even \oursS outperforms \siglipBmlf by 9–10\% and \siglipBsf by 22–24\%. To analyze this effect, Table~\ref{tab:cc3m-aug} presents a breakdown of performance across individual tasks in \cciiiaug. %
The advantage of \ours emerges specifically in tasks requiring identification of the correct image after cropping, rotation, or flipping (horizontal or vertical). 
Unlike the baselines, \ours solves these tasks almost perfectly (see qualitative examples in Figs.~\ref{fig:cc3m-aug-eval-short} and \ref{fig:cc3m-aug-eval}).
\revMAoJ{
We note that TGIT is evaluated on a held-out evaluation set, but the same transformation families are also part of the training mixture. Thus, these results measure the learnability of text-guided transformations rather than generalization to unseen transformation families.
}%
The improvement over baselines likely stems from the nature of these tasks, which rely on capturing the visual structure rather than semantic content.
The unimodal encoders tend to extract semantic information which is used to align different inputs in the latent space. It is reasonable to hypothesize that features at deeper layers contain higher amount of semantic information, at the expense of other aspects such as visual information~\chst{\cite{dorszewski2025colors}}.
Moreover, attending to both image and text is crucial to solve these tasks, as no shortcut can be learned by just looking at input images and target images or input text and target images. Late fusion models likely have limited access to the information necessary to solve the task, while our early fusion approach can easily learn it.
\revHYHX{%
This is further supported by the comparison between late fusion baselines: SigLIP\textsubscript{MLF}, which uses a learnable fusion network, substantially outperforms SigLIP\textsubscript{SF} on 3 out of 5 transformations, while achieving comparable performance on the remaining ones. Thus, learnable late fusion can improve over simple score fusion in several cases, but remains limited compared to early fusion.
}%
This interpretation can also explain the better results of \ours compared to the baselines on \oipos, where the model needs to distinguish left and right instances of the same object.

\textbf{Fine-tuning on MMEB.}
We finetune all models that were trained on \ccxiimm on the training split of the MMEB dataset~\cite{jiang2025vlm2vec} for 10 epochs and report the results in Table~\ref{tab:finetune-mmeb}.
All models improve on the MMEB tasks \textit{Classification, VQA, and Retrieval}, while suffering from performance drops in \textit{Grounding}, which is likely due to the training set holding only few grounding samples.
Models improve on \imnet since it is part of the MMEB training split. Moreover, we observe unlearning on the \vgcrop, \oicrop, and \cciiiaug tasks.

\textbf{Comparison to other models.}
We compare to embedding models from other works in Table~\ref{tab:finetune-mmeb}, namely a pre-trained \siglip model \cite{zhai2023sigmoid} and VLM2Vec \cite{jiang2025vlm2vec}.
We evaluate pre-trained \siglip on multimodal embedding tasks via score fusion. This model has undergone much longer pre-training than our models, thus attains high performance on tasks that mainly require unimodal embeddings, such as Classification and \imnet. However, our models are better on multimodal tasks. Note that the MMEB results of the pre-trained \siglip are higher than reported in the online leaderboard,\footnote{\scriptsize \url{https://huggingface.co/spaces/TIGER-Lab/MMEB-Leaderboard}} since we use more suitable prompts as described in App.~\ref{app:eval-tasks}. 
Moreover, we evaluate a VLM2Vec model \cite{jiang2025vlm2vec} that is based on the Phi-3.5-V large vision-language model and fine-tuned on MMEB. Notably, this model has much more parameters (4.15B, i.e.~over 25x that of \oursB) and has undergone much longer pre-training. Expectedly, this model outperforms our MMEB-fine-tuned models on MMEB evaluation tasks. However, both \oursS and \oursB achieve higher performance on \revMAoJ{our} \vgcrop, \oicrop, \oipos, and \cciiiaug{}  \revMAoJ{tasks}, thus highlighting the effectiveness of early fusion on \revMAoJ{our} challenging multimodal tasks. Both pre-trained \siglip and VLM2Vec perform below random chance level on \cciiiaug, due to their bias towards selecting the original image rather than the transformed one from the retrieval pool.

\begin{table*}[t]
\small
\caption{%
\textbf{Fine-tuning on MMEB and comparison to other models.} We report evaluation results for models fine-tuned on the training split of MMEB~\cite{jiang2025vlm2vec}, using as initialization the models that were trained on \ccxiimm. Moreover, we compare to models from other works: pretrained \siglip~\cite{zhai2023sigmoid} (much longer pretraining than our models) and VLM2Vec \cite{jiang2025vlm2vec} (much larger model and much longer pretraining than our models).
}
\centering
\tabcolsep=1.7pt
\resizebox{\linewidth}{!}{
\begin{tabular}{l@{\hspace{6pt}}l ccccccccc
}
    \toprule
    Train Data & Model & Classif. & VQA & Retrieval & Grounding & ImNet & VG-Crop & OI-Crop & OI-Pos & TGIT \\
    \midrule
    & \siglipSsf & 38.4 & 20.5 & 32.2 & 55.3 & 31.1 & 36.2 & 39.5 & 44.3 & 41.6 \\
    & \siglipSmlf & 38.0 & 21.6 & 31.9 & 61.3 & 37.3 & 33.9 & 50.5 & 44.3 & 43.6 \\
    & \siglipBsf & 40.4 & 22.0 & 34.6 & 55.1 & 36.6 & 36.7 & 41.7 & 45.4 & 43.7 \\
    & \siglipBmlf & 39.8 & 22.1 & 34.1 & 61.9 & 41.8 & 36.4 & 47.5 & 43.0 & 39.9 \\
    \cmidrule(l){3-11}
    \rowcolor{myBlue} \cellcolor{white}  & \oursS & 33.6 & 22.6 & 27.2 & 62.8 & 33.2 & 32.4 & 47.0 & 52.9 & 50.9 \\
    \rowcolor{myBlue} \cellcolor{white} \multirow{-7}{*}{\makecell{\ccxii\\+MM\\$\rightarrow$MMEB}} & \oursB & 40.4 & 24.9 & 34.9 & 73.3 & 41.9 & 36.4 & 51.8 & 53.1 & 54.8 \\
    \midrule
    Pretrained & \siglip-B/16~{\scriptsize\cite{zhai2023sigmoid}} & 51.3 & 11.0 & 47.1 & 58.2 & 76.1 & 8.1 & 27.6 & 34.2 & 6.6 \\
    \midrule
    \makecell{Pretrained\\$\rightarrow$MMEB} \hfill & \makecell[l]{VLM2Vec\\Phi-3.5-V}~{\scriptsize\cite{jiang2025vlm2vec}} & 53.1 & 55.1 & 63.4 & 77.3 & 60.0 & 15.5 & 37.7 & 50.0 & 12.4 \\
    \midrule
     \hspace{5mm} -- & Random Chance & 5.6 & 0.1 & 0.1 & 0.1 & 0.1 & 6.3 & 10.0 & 26.6 & 22.0 \\
    \bottomrule
\end{tabular}}
\label{tab:finetune-mmeb}
\end{table*}

\begin{table*}[t]
\small
\centering
\tabcolsep=1.8pt
\caption{%
\textbf{Ablations.} We report the effect of removing hard negatives (Sec.~\ref{sec:training-scheme}) or the MMM loss (Sec.~\ref{sec:training-fuselip}) when training on \cciiimm. Hard negatives are crucial for performance on \vgcrop, \oicrop, \oipos, TGIT. The MMM loss can only be applied to \ours and improves performance across almost all tasks.
}
\begin{tabular}{L{18mm} | cc | ccccccccc
}
    \toprule

    Model & Hard Neg. & $\L_\textrm{MMM}$ & Classif.\zs & VQA\zs & Retrieval\zs & Grounding\zs & ImNet\zs & VG-Crop & OI-Crop\zs & OI-Pos\zs & TGIT \\
    \midrule
    & \xmark & - & 19.5 & 13.2 & 11.3 & 80.6 & 8.6 & 23.4 & 39.1 & 40.5 & 11.5 \\
    \multirow{-2}{*}{\siglipSsf} & \cmark & - & 21.5 & 12.7 & 13.0 & 74.8 & 8.8 & 52.0 & 55.2 & 45.4 & 57.3  \\
    \midrule
    & \xmark & - & 17.1 & 13.5 & 11.4 & 76.5 & 9.7 & 34.1 & 46.6 & 43.0 & 22.1 \\
    \multirow{-2}{*}{\siglipSmlf} & \cmark & - & 18.0 & 14.2 & 12.7 & 74.2 & 10.2 & 53.0 & 66.2 & 46.9 & 67.2 \\
    \midrule
     & \xmark & \cmark & 20.0 & 16.5 & 11.9 & 76.2 & 15.1 & 35.0 & 51.0 & 51.7 & 18.5 \\
    & \cmark &  \xmark & 17.8 & 15.7 & 10.5 & 66.2 & 12.3 & 46.6 & 58.8 & 50.7 & 83.8 \\
    \rowcolor{myBlue} \cellcolor{white} \multirow{-3}{*}{\oursS} &\cmark & \cmark & 18.5 & 15.9 & 11.2 & 70.8 & 13.5 & 49.6 & 59.8 & 53.9 & 79.0 \\
    \midrule
    & \xmark & \cmark & 23.3 & 17.5 & 14.6 & 82.5 & 18.4 & 38.8 & 52.5 & 45.3 & 13.6 \\
    & \cmark &  \xmark & 21.0 & 16.8 & 14.3 & 81.3 & 16.3 & 53.7 & 63.3 & 68.0 & 88.4 \\
    \rowcolor{myBlue} \cellcolor{white} \multirow{-3}{*}{\oursB} &\cmark & \cmark & 23.3 & 17.5 & 15.0 & 82.4 & 18.1 & 55.8 & 68.1 & 70.8 & 94.3 \\
    \bottomrule
\end{tabular}
\label{tab:ablation_fuse-clip}
\end{table*}

\subsection{Ablation on the importance of hard negatives and masked modeling loss}
\label{sec:ablation}

To better understand the contribution of some design choices in \ours, we train models selectively removing the hard negatives from training (Sec.~\ref{sec:training-scheme}) and the masked modeling loss. Results are reported in Table~\ref{tab:ablation_fuse-clip}.
First, we see that not including hard negatives in the batch causes large drops in performance on \vgcrop, \oicrop, \oipos, and especially \cciiiaug (e.g., for \oursB accuracy decreases from 94.3\% to 13.6\%). Moreover, adding the hard negatives does not affect performance on the other tasks for \oursB, and yields only a minor decrease for the smaller \oursS.
Similar observations about the benefit of hard negatives hold true also for the late fusion baselines.
Second, Table~\ref{tab:ablation_fuse-clip} illustrates the role of the MMM loss: 
\revXRX{
it yields improvements across most tasks, in particular for the larger \oursB.
However, even without the MMM loss \oursB remains substantially stronger than the baselines on most tasks, thereby confirming that it is not the sole driver of \ours's gains.
Rather, the MMM loss provides an additional improvement that is naturally supported by the shared discrete-token architecture. %
}%
We provide additional evidence of the advantages given by the MMM loss in Table~\ref{tab:ablation_mlm_extended} in the Appendix.

Further ablations and properties of \ours such as variance over random seeds for training, using discrete image tokens in late fusion, compositionality, and the modality gap are studied in Appendix~\ref{app:additional-results}.

\section{Discussion}

The results of \ours have several implications. First, it is possible to train a CLIP-like model (on unimodal or multimodal data) using a single encoder, in contrast to standard \clip models that rely on separate text and image encoders.
Second, our architecture, which inherently supports multimodal embeddings, enables seamless integration of contrastive and masked modeling objectives.
This significantly simplifies the FLAVA training setup (see Sec.~\ref{sec:training-fuselip}), showing that both objectives can be combined without requiring separate forward passes. Moreover, our models achieve stable training
by using standard recipes.
Third, our results highlight tasks, such as the text-guided transformations, where early fusion of modalities  significantly outperforms late fusion. Since solving such tasks is part of a comprehensive multimodal encoder, we argue that early fusion is particularly promising for multimodal embeddings.
Finally, we anticipate that \ours can be naturally extended to new applications, e.g., encoding multiple images.%

\textbf{Limitations.}
Given limited computational resources, we could not test the effect of further scaling data and model size. Moreover, while our models are faster and require significantly less memory at training time, they are more expensive at inference than the baselines, but we expect the gap to be reduced at scale and thanks to the ongoing progress of image tokenizers.

\section{Conclusion}
We have introduced a novel approach to multimodal embedding models, designing an early fusion architecture with discrete image tokenization and a single encoder.
Our simple training recipe 
combines contrastive and masked modeling objectives, while leveraging hard negative samples.
The individual components as well as the final models are empirically \mbox{validated} on a variety of tasks. \chst{We demonstrate the benefits of early fusion in a controlled head-to-head comparison against late fusion baselines.}
Our methods for generating training datasets that are tailored for multimodal learning can be applied at scale.
Moreover, our novel evaluation tasks are complementary to the existing benchmarks typically used for testing embedding models.
Overall, we believe our approach and datasets can be valuable building blocks for future research on multimodal embedding models.

\section*{Broader Impact}
\revMAoJ{
This work is methodological and aims to improve multimodal representation learning. It inherits potential societal risks from its training data setup. Models trained on \cciii, \ccxii, and OpenImages may reproduce biases, stereotypes, and coverage imbalances present in these web-scale datasets. Moreover, our \cciiivqa data is generated from captions using and LLM, and can therefore encode both caption biases and language-model priors. Finally, capabilities such as solving \oipos, which require distinguishing spatially localized objects may be relevant to applications with privacy concerns such as identification, localization, or tracking, when combined with other systems.
}%
\section*{Acknowledgements}  
We thank the International Max Planck Research School for Intelligent Systems (IMPRS-IS) for supporting C.S. We acknowledge support from the Deutsche Forschungsgemeinschaft (DFG, German Research Foundation) under Germany’s Excellence Strategy (EXC number 2064/1, project number 390727645), as well as in the priority program SPP 2298, project number 464101476. 
F.C. and N.F. acknowledge support from an unrestricted gift from Google and the Swiss National Science Foundation (grant number 212111). 
Any opinions, findings, and conclusions or recommendations expressed in this material are those of the authors and do not necessarily reflect the views of the sponsors.

\bibliographystyle{plainnat}
\bibliography{main}

\clearpage
\appendix

\onecolumn  %

\section*{Appendix}

This appendix provides additional details and results to support the main text. In App.~\ref{app:implementation-details} we report details on the data creation, model implementation, training scheme, and evaluation tasks. In App.~\ref{app:additional-results} we proceed to report complementary results, specifically regarding compositionality, the modality gap, training late fusion baselines with frozen encoders, training on unimodal data, and further ablations.

\section{Implementation Details}
\label{app:implementation-details}

\subsection{Data}
\label{app:data}

\myparagraph{\cciiiaug.}
In the following we give a detailed description of the transformations applied in \cciiiaug and \ccxiiaug.
\textit{random cropping}: crop to a location of the image, specified as ``upper left, upper center,\ldots'', thus there are 9 possible crop locations. \textit{random rotations}: randomly rotate the image clockwise or counter-clockwise at a random angle sampled uniformly from (10°, 20°,\ldots, 90°). \textit{flipping}: flip the image vertically or horizontally. \textit{colorization}: convert the image to grayscale and use the original colored image as the target. \textit{grayscale}: convert the image to grayscale and use it as the target, use the original image as query. \textit{jitter}: apply color-jittering by randomly adjusting brightness, contrast, and saturation by random factors, which are sampled uniformly between 0.3 and 2.0 and rounded to one decimal place.

\myparagraph{\cciiivqa.}
We generate VQA data from \cciii by querying an LLM to rewrite the given captions into question-answer pairs. To this end, we use Llama-3.1-8B-Instruct~\cite{dubey2024llama} with a custom system prompt that guides the model to produce QA-pairs via rules and examples. The full system prompt is reported in Fig.~\ref{fig:system_prompt}. Notably, the model receives only the captions (not the images).
In order to encourage variability in the generated VQA data, we use stochastic decoding by sampling the next token based on the predicted probabilities.

\subsection{Models}
\label{app:model-details}

Following \siglip~\cite{zhai2023sigmoid}, we use bidirectional attention for all models.
Moreover, we generally use a context length of 180 and mask out padding tokens. The baseline models are based on the \vits and \vitb \clip models as implemented in OpenCLIP~\cite{cherti2023openclip}. We present links and licenses for all assets used in this paper in Table~\ref{tab:assets-licenses}.

\myparagraph{\siglipsf.}
For the score-level fusion of \siglip, we simply add the normalized features arithmetically, and normalize again after the addition.

\begin{wraptable}{r}{0.55\textwidth}
\centering
\small
\caption{\textbf{Memory and runtime comparison.}}
\tabcolsep=1.5pt
\begin{tabular}{l@{\hspace{-3pt}} *{4}{c}}
    \toprule
    & \multicolumn{2}{c}{Training} & \multicolumn{2}{c}{Inference} \\
    \cmidrule(l{3pt}r{3pt}){2-3} \cmidrule(l{3pt}r{3pt}){4-5}
    & Memory \tiny [GB] & Time \tiny [ms] & Memory \tiny [GB] & Time \tiny [ms] \\
    \midrule
    \siglipSsf  & 18.7 & 315   & 0.7 & 100 \\
    \siglipSmlf  & 18.8 & 328   & 0.7 & 104 \\
    \siglipBsf   & 32.2 & 595  & 1.5 & 199 \\
    \siglipBmlf  & 32.3 & 612   & 1.5 & 202 \\
    \midrule
    \oursS   & 11.0 & 243  & 1.2 & 129 \\
    \oursB   & 14.1 & 425  & 2.3 & 271 \\
    \bottomrule
\end{tabular}
\label{tab:runtime}
\end{wraptable}

\myparagraph{\siglipmlf.}
Following the original implementation of \magiclens~\cite{zhang2024magiclens}, the fusion module of \siglipSmlf is a transformer, for which we scale the width and number of heads down to match those of \vits. Thereby, the fusion module has 4 layers at width 384 with 6 heads. For \mbox{\siglipBmlf} the fusion module has 4 layers at width 512 with 8 heads. The fusion module operates on the concatenated non-normalized image and text embeddings and is followed by an attention pooling layer. Whenever the model is queried without an image, we use a zero-tensor for the image embedding in the late fusion stage.

\myparagraph{\ours.}
For \oursS we use the TiTok family of image tokenizers~\cite{yu2024titok}. In particular, for \oursS we use TiTok-S-128 and the subsequent transformer is based on the text transformer of S-sized \clip (width of 384, 6 heads, 12 layers). For \oursB, we use the TiTok-BL-128-VQ image tokenizer~\cite{yu2024titok} followed by the text transformer of B-sized \clip (width of 512, 8 heads, 12 layers). We do not use the generators that were trained along with the image tokenizers.

\myparagraph{Memory requirements and runtime.}
In Table~\ref{tab:runtime} we show a comparison of the memory allocation and runtime at training and at inference time of the models considered in this paper. To this end we do forward passes at batch size 128 of two multimodal samples (image+text) and average this over 10 repetitions. The \ours models profit at training time from fewer trainable parameters and exhibit significantly lower memory allocation than the baselines. This means that our models can be trained with the same batch size on fewer GPUs and thus save resources.
As a drawback, the forward pass of \ours is more expensive at inference since images are processed sequentially by both tokenizer and encoder. However, we expect this overhead to be reduced when scaling up the size of the encoder, which will dominate inference cost.
\revMAoJ{
For retrieval applications, embeddings of target images can typically be precomputed during indexing, so the tokenizer and encoder cost for the retrieval database is amortized.
}%

\begin{figure*}[t]
    \centering
    \begin{minipage}{0.97\textwidth}
    \centering
    \lstset{
        backgroundcolor=\color{gray!10}, %
        basicstyle=\ttfamily\footnotesize,      %
        frame=single,                    %
        breaklines=true                   %
    }
\begin{lstlisting}
You are a helpful assistant, that generates a question and answer about an image that has a given caption. Rules:
1. For generating question/answer pairs, only use information that is evident from the caption.
2. Do not mention the word 'caption' in the question or answer.
3. Answers should be at least a couple words long (not single word).
4. Don't start every question with "What"
5. Respond in the format: Question: <question>
Answer: <answer>
Examples:
Caption: A group of friends are having a barbecue in the backyard. 
Question: Where is the barbecue taking place?
Answer: In the backyard.
Caption: A child is playing with a toy airplane on the floor. 
Question: Which toy is the child holding?
Answer: A toy airplane.
Caption: A girl with a smartphone is lying on the sofa. 
Question: Where is the girl in the image located?
Answer: On the sofa.
Caption: A golden retriever is running through a field of flowers. 
Question: What is the dog doing?
Answer: Running through a field of flowers.
\end{lstlisting}
    \end{minipage}
    \caption{\textbf{System prompt for generating question-answer pairs.} We use Llama-3.1-8B-Instruct to generate VQA samples from image-text pairs of \cciii as described in Sec.~\ref{sec:multimodal-data}.
    }
    \label{fig:system_prompt}
\end{figure*}

\subsection{Training}
\label{app:training}

\begin{wraptable}{r}{0.55\columnwidth}
\centering
\small
\vspace{-12mm}
\caption{\textbf{Hyperparameters for training.}}
\begin{tabular}{lll}
    \toprule
    Parameter & \cciiimm & \ccxiimm \\
    \midrule
    Epochs & 8 & 16 \\
    Optimizer & AdamW & AdamW \\
    Batch size & 2048 & 2048 \\
    Weight decay & 1.0 & 0.5 \\
    AdamW $\beta_1, \beta_2$ & 0.9, 0.98 & 0.9, 0.98 \\
    AdamW $\epsilon$ & $1 \times 10^{-8}$ & $1 \times 10^{-8}$ \\
    Learning rate & $1 \times 10^{-3}$ & $1 \times 10^{-3}$ \\
    Learning rate schedule & cosine & cosine\\
    Warmup steps & 12000 & 12000 \\
    $\ell_2$-gradient clipping & 1.0 & 1.0 \\
    Context length & 180 & 180 \\
    Image resolution & 256 & 256 \\
    \bottomrule
\end{tabular}
\label{tab:hyperparams-training}
\end{wraptable}

\textbf{Batch composition.}
We make sure that hard negatives are present in each batch by the sampling strategy outlined in Sec.~\ref{sec:training-scheme}. Here we give a detailed account of the amount of hard negatives sampled for each
transformation in \cciiiaug and \ccxiiaug. \textit{crop}: we take all 9 possible crops. \textit{rotate}: we take 3 randomly selected rotations. \textit{jitter}: we take 3 samples. \textit{flip}: we take horizontal flip and vertical flip, both applied once from the original image and once from the flipped image, yielding 4 samples. \textit{colorize-grayscale}: for each colorization sample we also take the to-grayscale sample, and vice-versa, yielding 2 samples.

\textbf{Hyperparameters.}
The hyperparameters used for training on \cciiimm and \ccxiimm are reported in Table~\ref{tab:hyperparams-training}. We use $\ell_2$-norm gradient clipping for all models as we observed that this stabilizes multimodal training. We select AdamW~\cite{loshchilov2018decoupled} as the optimizer.

\subsection{Evaluation tasks}
\label{app:eval-tasks}

\myparagraph{Massive Multimodal Embedding Benchmark (MMEB).}
MMEB~\cite{jiang2025vlm2vec} is a multimodal embedding benchmark consisting of 36 datasets that are split into the categories Classification, VQA, Retrieval, and Grounding. Each dataset consists of 1000 samples, and for each sample there are 1000 candidates, except for classification datasets that contain less classes. Only one of the candidates is correct.
Since MMEB was created to train and evaluate \vlmtovec \cite{jiang2025vlm2vec}, an embedding model that is based on an autoregressive large multimodal model, it contains prompts in instruction format. As our models are not trained on instruction-following data, we remove this part of the prompts (except when evaluating the VLM2Vec model in Table~\ref{tab:finetune-mmeb}).

\myparagraph{OpenImages-Crop (\oicrop.)}
To create the \oicrop task, we start with the OpenImages dataset and first remove all bounding boxes that are very small (less than 50 pixels), very large (larger than 0.9 relative size in either dimension), that have high aspect ratio (larger than 1.5). Then we remove boxes with labels that appear less than ten times in total. Next, we filter out samples that have less than five uniquely labelled bounding boxes, then we drop for each sample bounding boxes that are strongly overlapping (IoU larger than 0.6). Finally, we gather for each sample a label with the corresponding bounding box, and collect as negatives four bounding boxes from the same image (with different label) and five bounding boxes from other images with the same label. Since the distribution of label names is heavily skewed, we make sure to get each label name at maximum five times as a query.
In total this procedure yields 1046 samples, some are shown in Fig.~\ref{fig:oi-vg-crop-tasks}.

\myparagraph{OpenImages-Position (\oipos).}
For \oipos we select OpenImages images that contain an object exactly twice. We consider 34 object classes, chosen so that identification of individual instances is generally possible (e.g. for ``Window'' this is not the case, as many images with windows contain additional windows that are not labelled). We drop images where the bounding boxes overlap significantly in horizontal direction (if the horizontal center of either box overlaps the other box). Next, we obtain one negative bounding box each from the left and right border of the image. If this would cause overlap to the positive box, we obtain boxes from the top or bottom. If there is also not enough space (we enforce min.~size of 30x30 pixels), the retrieval pool could contain less than four samples. Then we restrict each label to 100 samples. The query is then the image with the text ``The \{object\_name\} on the left/right'', and the retrieval pool consists of the two object crops and the two border crops. In total this task contains 2546 samples. See Fig.~\ref{fig:oi-vg-crop-tasks} for some examples.

\myparagraph{\imnet.}
We evaluate on the full \imnet-1k validation set~\cite{deng2009imagenet}, and use the ensemble of OpenAI prompt templates~\cite{radford2021clip}.
In contrast, the \imnet evaluation as part of MMEB uses only a single prompt and 1k samples.

\begin{table*}[t]
    \centering
    \caption{\textbf{Assets used in this paper.}}
    \small
    \begin{tabular}{lll}
        \toprule
         Asset & Link & License  \\
         \midrule
         OpenCLIP & \scriptsize \url{https://github.com/mlfoundations/open_clip} & MIT \\
         TiTok & \scriptsize \url{https://github.com/bytedance/1d-tokenizer} & Apache-2.0 \\
         Llama-3.1-8B-Instruct & \scriptsize \url{https://huggingface.co/meta-llama/Llama-3.1-8B-Instruct} & see Link \\
         VLM2Vec & \scriptsize \url{https://huggingface.co/TIGER-Lab/VLM2Vec-Full} & Apache 2.0 \\
         \cciii & \scriptsize \url{https://huggingface.co/datasets/pixparse/cc3m-wds} & see Link \\
         \ccxii & \scriptsize \url{https://huggingface.co/datasets/pixparse/cc12m-wds} & see Link \\
         MMEB & \scriptsize \url{https://huggingface.co/datasets/TIGER-Lab/MMEB-eval} & Apache 2.0 \\
         VisualGenome & \scriptsize \url{https://homes.cs.washington.edu/~ranjay/visualgenome/index.html} & CC BY 4.0 \\
         OpenImages & \scriptsize  \url{https://storage.googleapis.com/openimages/web/index.html} & CC BY 4.0 \\
         \scp & \scriptsize  \url{https://github.com/RAIVNLab/sugar-crepe} & MIT \\
         \bottomrule
    \end{tabular}
    \label{tab:assets-licenses}
\end{table*}

\section{Additional Results}
\label{app:additional-results}

\begin{table}
\small
\newcommand{\std}[1]{{\scriptsize$\pm$#1}}
\caption{\revMAoJ{
\textbf{Variance of training runs.} We report mean and standard deviation over 3 training runs on \cciiimm with different random seeds. \ours maintains a clear advantage over \siglipBmlf on most benchmarks. \emph{\zs{}zero-shot}
}}
\centering
\setlength{\tabcolsep}{5pt}
\resizebox{\linewidth}{!}{%
\begin{tabular}{lrrrrrrrrr}
\toprule
& Classif.\zs & VQA\zs & Retrieval\zs & Grounding\zs & \imnet\zs & VG-Crop & OI-Crop\zs & OI-Pos\zs & TGIT \\
\midrule
\siglipBmlf & 20.2\std{0.8} & 14.3\std{0.4} & 14.0\std{0.2} & 75.7\std{1.0} & 12.3\std{0.4} & \textbf{55.9}\std{0.8} & \textbf{67.8}\std{0.8} & 47.5\std{0.2} & 70.9\std{1.7} \\
\rowcolor{myBlue} \oursB & \textbf{23.0}\std{0.5} & \textbf{17.2}\std{0.5} & \textbf{15.0}\std{0.8} & \textbf{82.1}\std{0.6} & \textbf{17.2}\std{1.5} & \textbf{55.6}\std{1.9} & \textbf{67.4}\std{1.2} & \textbf{69.0}\std{1.5} & \textbf{91.2}\std{5.5} \\
\bottomrule
\end{tabular}%
}
\label{tab:results-multi-seed}
\end{table}

\begin{table*}
\small
\caption{%
\textbf{\textbf{Evaluation on \scp.}}
\ours{} outperforms the baselines on most compositionality tasks.
}
\centering
\setlength{\tabcolsep}{2pt}
\begin{tabular}{llcccccccccc}
    \toprule
    \multirow{2}{*}{Train Data} & \multirow{2}{*}{Model} &\multicolumn{4}{c}{Replace} & \multicolumn{3}{c}{Swap} & \multicolumn{3}{c}{Add} \\
    \cmidrule(lr){3-6} \cmidrule(lr){7-9} \cmidrule(lr){10-12}
     &  & Object & Attribute & Relation & \textbf{Mean} & Object & Attribute & \textbf{Mean} & Object & Attribute & \textbf{Mean} \\
    \midrule

    & \siglipSsf & 67.4 & 69.0 & 56.8 & 64.4 & 51.8 & 54.5 & 53.2 & 46.5 & 46.1 & 46.3 \\
    & \siglipSmlf & 71.8 & 69.8 & \textbf{59.9} & 67.2 & 52.2 & 59.5 & 55.8 & 66.8 & 60.1 & 63.4 \\
    & \siglipBsf & 70.4 & 71.2 & 59.5 & 67.0 & 47.4 & 58.4 & 52.9 & 48.2 & 46.1 & 47.1 \\
    & \siglipBmlf & 75.8 & 70.7 & 59.5 & 68.7 & \textbf{58.0} & 59.3 & 58.6 & 69.7 & 62.3 & 66.0 \\
    \cmidrule(l){3-12}
    \rowcolor{myBlue} \cellcolor{white} & \oursS & 74.2 & 69.5 & 58.8 & 67.5 & 50.6 & 62.9 & 56.8 & 71.3 & 60.8 & 66.1 \\
    \rowcolor{myBlue} \cellcolor{white} \multirow{-7}{*}{\makecell{\cciii\\+\mmdata}} & \oursB& \textbf{78.8} & \textbf{71.3} & 59.4 & \textbf{69.8} & 56.3 & \textbf{64.0} & \textbf{60.1} & \textbf{73.8} & \textbf{66.8} & \textbf{70.3} \\
    \midrule

    & \siglipSsf  & 79.6 & 74.4 & 65.1 & 73.0 & 62.0 & 63.7 & 62.8 & 55.9 & 55.4 & 55.6 \\
    & \siglipSmlf & 85.4 & 78.3 & \textbf{69.7} & 77.8 & 62.4 & 65.8 & 64.1 & 78.8 & 72.4 & 75.6 \\
    & \siglipBsf & 82.4 & 76.0 & 64.2 & 74.2 & 63.3 & 63.5 & 63.4 & 54.4 & 52.0 & 53.2 \\
    & \siglipBmlf & \textbf{88.3} & 78.3 & 67.8 & 78.1 & \textbf{66.5} & 69.4 & 68.0 & 79.0 & \textbf{73.4} & 76.2 \\
    \cmidrule(l){3-12}
    \rowcolor{myBlue} \cellcolor{white} & \oursS & 83.4 & 77.4 & 66.4 & 75.8 & 60.0 & 68.0 & 64.0 & 76.6 & 71.0 & 73.8 \\
    \rowcolor{myBlue} \cellcolor{white} \multirow{-7}{*}{\makecell{\ccxii\\+\mmdata}} & \oursB & 87.6 & \textbf{78.9} & 69.6 & \textbf{78.7} & \textbf{66.5} & \textbf{69.8} & \textbf{68.2} & \textbf{82.4} & 72.0 & \textbf{77.2} \\
    \bottomrule
\end{tabular}
\label{tab:scp-app}
\end{table*}

\myparagraph{Variance of training runs.}
\revMAoJ{
To assess the stability of our results, we repeat training of \siglipBmlf and \oursB on \cciiimm with 3 different random seeds, reporting mean and standard deviation in Table~\ref{tab:results-multi-seed}. \oursB outperforms \siglipBmlf on most benchmarks, with gains that exceed the observed run-to-run variation. The only exceptions are \vgcrop and \oicrop, where both methods perform comparably. 
}%

\myparagraph{Compositionality.}
The \scp benchmark~\cite{hsieh2023sugarcrepe} measures how well models capture compositional concepts in text and images. This benchmark considers modifying objects, attributes, and relations of sentences via replacing, swapping, and adding.
To test how the different approaches to
multimodal embeddings influence the compositionality ability, we report their performance on \scp in \mbox{Table}~\ref{tab:scp-app}. 
In both training setups (\cciiimm and \ccxiimm), our \oursB attains the highest mean performances across compositionality categories.
This result suggests another potential benefit of early fusion.

\myparagraph{Modality gap.}
The modality gap in vision-language models with unimodal embedding, i.e. images and text are mapped to different regions of the latent space, is a well-known phenomenon~\cite{liang2022mind,shi2023towards}.
With multimodal embedding models we can also study the relative position of the representations of multimodal inputs.
In Table~\ref{tab:modality-gap} we compute the modality gap as defined by \citet{liang2022mind}:
\begin{align*}
    \norm{\frac1n \sum_{i=1}^n f(z^1_i) - \frac1n \sum_{i=1}^n f(z^2_i)}_2,
\end{align*}
where $z^1_i$ and $z^2_i$ are inputs from different modalities.
We select datasets representing the four combinations of data modalities spanned by the evaluation tasks. These datasets are sub-tasks of MMEB~\cite{jiang2025vlm2vec}.
We observe that the results are similar across fusion models.
\begin{wraptable}{r}{0.55\textwidth}
\centering
\small
\caption{\textbf{Modality gap.} We report the modality gap (as defined by~\citet{liang2022mind}) for models at initialization, and trained on \ccxiimm.}
\setlength{\tabcolsep}{2pt}
\begin{tabular}{llcccc}
    \toprule
    && ImNet & OK-VQA  & RefCOCO & RefCOCO-M \\
    \cmidrule{3-6}
    && I -- T & IT -- T & IT -- I & IT -- IT \\
    \midrule
    &\siglipSsf & 1.24 & 0.86 & 0.65 & 0.02 \\
    &\siglipSmlf & 1.04 & 0.97 & 0.71 & 0.02 \\
    \multirow{-3}{*}{\rotatebox{90}{init.}} &\oursS & 0.72 & 0.68 & 0.19 & 0.01 \\
    \midrule
    &\siglipSsf & 0.44 & 0.59 & 0.69 & 0.02 \\
    &\siglipSmlf & 0.39 & 0.46 & 0.64 & 0.02 \\
    \multirow{-3}{*}{\rotatebox{90}{trained}} &\oursS & 0.42 & 0.52 & 0.58 & 0.02 \\
     \bottomrule
\end{tabular}
\label{tab:modality-gap}
\end{wraptable}
Interestingly, the gap between multimodal (IT) and unimodal samples (T or I) are larger than between unimodal inputs from different modalities (I -- T), while the gap for IT -- IT data is very small as expected.
Finally, we test the modality gap at initialization, i.e., with random encoders: early fusion (\oursS), with more shared weights, yields smaller gaps than late fusion, but this difference disappears after training.

\begin{table*}[!t]
\small
\caption{%
\chst{\textbf{Ablation of the image tokenizer and transformer sizes.} We vary the image tokenizer and transformer sizes for models trained on \cciiimm. While scaling the image tokenizer drives performance, scaling up the transformer also yields further improvements. The highlighted rows correspond to \oursS and \oursB respectively. \emph{\zs{}zero-shot}}
}
\centering
\tabcolsep=2.5pt
\begin{tabular}{*{11}{c}
}
\toprule
Tokenizer & Transformer & Classif.\zs & VQA\zs & Retrieval\zs & Grounding\zs & \imnet\zs & VG-Crop & OI-Crop\zs & OI-Pos\zs & TGIT \\
\midrule
\rowcolor{myBlue} S & S & 18.5 & 15.9 & 11.2 & 70.8 & 13.5 & 49.6 & 59.9 & 53.9 & 78.5 \\
S & B & 19.4 & 16.4 & 12.0 & 72.7 & 14.6 & 48.3 & 58.4 & 55.5 & 91.5 \\
B & S & 22.1 & 16.6 & 14.3 & 81.6 & 16.1 & 55.4 & 66.2 & 68.0 & 95.6 \\
\rowcolor{myBlue} B & B & 23.3 & 17.5 & 15.1 & 82.5 & 18.1 & 55.8 & 68.1 & 70.8 & 93.9 \\
\bottomrule
\end{tabular}
\label{tab:tokenizer-ablation}
\end{table*}

\myparagraph{Ablation of the image tokenizer.}
We employ two image tokenizers in our work: \titok-S as part of \oursS, and \titok-B as part of \oursB. \citet{yu2024titok} report reconstruction FIDs on \imnet of 1.71 and 1.49 for \titok-S and \titok-B respectively, demonstrating higher quality of the larger tokenizer. In our experiments we indeed observe that this improves the embedding performance of FuseLIP. In order to further ablate the role of the tokenizer, we train \ours on \cciiimm in two new configurations: (i) with \titok-S and base-sized transformer encoder, and (ii) with \titok-B and small-sized transformer encoder. We report the results in Table~\ref{tab:tokenizer-ablation}. We observe that the higher quality \titok-B tokenizer already yields substantial improvements when used together with the small-sized transformer, confirming that tokenizer quality drives improvement. However, scaling up the transformer also contributes to improved performance for both tokenizers.
\revXRX{%
This ablation compares tokenizer fidelity at a fixed image-token budget, since both TiTok-S and TiTok-B represent each image with 128 tokens. Thus, the results indicate that higher tokenizer fidelity improves downstream performance without increasing the sequence length processed by the encoder. We do not vary the number of image tokens in this work, and therefore do not claim a scaling law with respect to token budget. However, increasing the token budget would directly increase the encoder sequence length and hence inference cost.
}%

\myparagraph{Late fusion of discrete image tokens.}
\revMAoJ{%
In order to disentangle the effect of the pretrained image tokenizer, we train a \siglipBmlf baseline, where the image encoder receives discrete image tokens from \titok-B. We compare this against \siglipBmlf and \oursB in Table~\ref{tab:titok-siglip-baseline}, trained on \cciiimm. We observe that this approach generally reduces performance compared to the standard \siglipBmlf baseline, thereby confirming that the pretraining of the image tokenizer is not the main driver of \ours performance, but rather its early fusion approach. 
}%

\chst{\textbf{Training \siglipmlf with frozen image and text encoders.}
For comparison, we test the performance of \siglipmlf when trained with frozen pre-trained image and text encoder backbones. In order to compare models trained at similar scale, we pre-train unimodal \siglip models on CC3M and use them as frozen initialization for \siglipmlf training on \cciiimm, which we denote as SigLIP-frozen\textsubscript{\textit{MLF}}. Results are reported in Table~\ref{tab:siglip-mlf-frozen-backbones}. Compared to \siglipmlf trained multimodally from scratch, we observe that the frozen initialization helps on some tasks (Classification, Retrieval), while significantly degrading performance on several others (Grounding, \vgcrop, \oicrop, \oipos, TGIT). We expect larger improvements when using a \siglip model pre-trained on larger scale data as initialization. However, such comparison would not be fair, as the pre-trained model would have undergone a significant amount of image-text alignment. In contrast, the image tokenizer used in FuseLIP has only been trained for image compression on \imnet (no textual alignment).}

\begin{table}
\small
\centering
\tabcolsep=3pt
\caption{\revMAoJ{
\textbf{Baseline with image tokenizer.} In order to disentangle the effect of the pretrained image tokenizer, we train on \cciiimm a \siglipBmlf baseline that receives image tokens from \titok-B. We observe that this generally reduces performance compared to the standard \siglipBmlf baseline, thereby confirming that the pretraining of the image tokenizer is not the main driver of \ours performance, but rather its early fusion approach. \emph{\zs{}zero-shot}
}}
\begin{tabular}{lrrrrrrrrr}
\toprule
 & Classif.\zs & VQA\zs & Retrieval\zs & Grounding\zs & \imnet\zs & VG-Crop & OI-Crop\zs & OI-Pos\zs & TGIT \\
\midrule
\siglipBmlf & 19.5 & 14.8 & 13.9 & 76.9 & 12.2 & 55.4 & \textbf{68.4} & 47.4 & 69.4 \\
\titok-\siglipBmlf & 18.6 & 14.2 & 11.8 & 68.2 & 12.1 & 51.1 & 58.8 & 47.3 & 69.3 \\
\rowcolor{myBlue} \oursB & \textbf{23.3} & \textbf{17.5} & \textbf{15.0} & \textbf{82.4} & \textbf{18.1} & \textbf{55.8} & 68.1 & \textbf{70.8} & \textbf{94.3} \\
\bottomrule
\end{tabular}
\label{tab:titok-siglip-baseline}
\end{table}

\begin{table*}
\small
\caption{%
\chst{\textbf{\siglipmlf with frozen backbones.}
We train \siglipmlf with frozen image and text encoder backbones (pre-trained on CC3M) on \cciiimm. These models are outperformed by the models trained from scratch on most tasks, on some significantly. \emph{\zs{}zero-shot}}
}
\centering
\tabcolsep=2.5pt
\resizebox{\textwidth}{!}{%
\begin{tabular}{l*{9}{c}
}
\toprule
Model & Classification\zs & VQA\zs & Retrieval\zs & Grounding\zs & \imnet\zs & VG-Crop & OI-Crop\zs & OI-Pos\zs & TGIT \\
\midrule
\siglipSmlf              & 18.0 & 14.2 & 12.7 & 74.2 & 10.2 & 53.0 & 66.2 & 46.9 & 67.4 \\
SigLIP-S-frozen\textsubscript{\textit{MLF}}  & 21.2 & 12.8 & 12.7 & 72.6 &  8.4 & 51.5 & 56.7 & 45.6 & 60.7 \\
\midrule
\siglipBmlf              & 19.5 & 14.8 & 13.9 & 76.9 & 12.2 & 55.4 & 68.4 & 47.4 & 69.4 \\
SigLIP-B-frozen\textsubscript{\textit{MLF}}  & 21.1 & 13.3 & 14.9 & 57.5 & 17.9 & 41.1 & 46.9 & 45.2 & 53.3 \\
\bottomrule
\end{tabular}}
\label{tab:siglip-mlf-frozen-backbones}
\end{table*}

\looseness=-1
\chst{\myparagraph{Loss ablation.}
As a sanity check, we report performance of \oursS trained on \cciiimm with only the masked multimodal modeling (MMM) loss in Table~\ref{tab:loss-ablation}. We observe that only using the MMM loss yields far worse performance, highlighting that the contrastive \siglip loss is necessary to properly train the \ours models.}

\begin{table*}
\small
\caption{%
\chst{\textbf{Loss ablation.}
We compare training \ours with only the masked multimodal modeling (MMM) loss. We observe that this yields weak performance, confirming that the employed contrastive loss is necessary.}
}
\centering
\tabcolsep=3pt
\resizebox{\textwidth}{!}{%
\begin{tabular}{l*{9}{c}
}
\toprule
 Model & Classification\zs & VQA\zs & Retrieval\zs & Grounding\zs & ImNet\zs & VG-Crop & OI-Crop\zs & OI-Pos\zs & TGIT \\
\midrule
FuseLIP-S-only-MMM  &  8.5 &  0.2 &  2.6 & 30.2 &  0.1 &  6.3 & 20.2 & 31.2 &  5.6 \\
\oursS              & 18.5 & 15.9 & 11.2 & 70.8 & 13.5 & 49.6 & 59.9 & 53.9 & 78.5 \\
\bottomrule
\end{tabular}}
\label{tab:loss-ablation}
\end{table*}

\chst{\myparagraph{Performance on unimodal tasks.}
While this paper focusses on multimodal embedding tasks, we also compare the performance of late and early fusion approaches on unimodal embedding tasks. These are tasks that do not require embedding image+text into a single feature vector, but rather into separate vectors, e.g. \imnet classification. The Classification and Retrieval subtasks of MMEB contain several tasks that are unimodal, in addition to our \imnet evaluation. In Table~\ref{tab:unimodal-tasks}, we report the mean performance of the models on these unimodal tasks. While \oursS trails some late-fusion baselines at \ccxiimm scale, our \oursB (same number of trainable parameters as SigLIP-S) consistently outperforms all baselines across tasks, indicating that early fusion does not sacrifice unimodal embedding ability.}

\begin{table*}
\small
\caption{%
\chst{\textbf{Unimodal tasks.}
We report average performances over unimodal tasks, split over \imnet, MMEB unimodal classification subtasks, and MMEB unimodal retrieval subtasks. \emph{\zs{}zero-shot}
}}
\centering
\tabcolsep=2.5pt
\begin{tabular}{ll *{3}{c}
}
\toprule
Train Data & Model & ImageNet\zs & Unimodal Classification\zs & Unimodal Retrieval\zs \\
\midrule
 & \siglipSsf  &  8.8 & 17.5 & 16.4 \\
 & \siglipSmlf & 10.2 & 13.8 & 16.3 \\
 & \siglipBsf  & 10.3 & 18.2 & 17.0 \\
 & \siglipBmlf & 12.2 & 15.8 & 18.1 \\
\rowcolor{myBlue} \cellcolor{white} & \oursS        & 13.5 & 14.3 & 14.3 \\
\rowcolor{myBlue} \cellcolor{white} \multirow{-6}{*}{\makecell{\cciii\\+\mmdata}} & \oursB        & 18.1 & 19.9 & 18.9 \\
\midrule
 & \siglipSsf  & 21.5 & 27.8 & 28.8 \\
 & \siglipSmlf & 25.5 & 27.4 & 28.9 \\
 & \siglipBsf  & 25.4 & 29.2 & 29.0 \\
 & \siglipBmlf & 28.8 & 28.1 & 28.1 \\
\rowcolor{myBlue} \cellcolor{white} & \oursS      & 26.0 & 22.6 & 25.3 \\
\rowcolor{myBlue} \cellcolor{white} \multirow{-6}{*}{\makecell{\ccxii\\+\mmdata}} & \oursB      & 32.7 & 29.3 & 31.7 \\
\bottomrule
\end{tabular}
\label{tab:unimodal-tasks}
\end{table*}

\begin{table*}
\small
\caption{%
\textbf{\cciii and \ccxii unimodal training.} We report evaluation results for models trained on \cciii and \ccxii without any multimodal embedding data.
}
\centering
\tabcolsep=1.5pt
\begin{tabular}{l@{\hspace{2pt}} l ccccccccc
}
    \toprule
    Train Data & Model & Classification\zs & VQA\zs & Retrieval\zs & Grounding\zs & ImageNet\zs & VG-Crop\zs & OI-Crop\zs & \oipos\zs & TGIT\zs \\
    \midrule
     &\siglip-S &  20.7 & 4.0 & 15.7 & 39.0 & 19.5 & 9.8 & 26.6 & 34.8 & 7.7 \\
    \rowcolor{myBlue} \cellcolor{white} & \oursS &  16.7 & 1.1 & 8.1 & 22.0 & 15.4 & 6.5 & 19.2 & 31.5 & 7.6 \\
    \rowcolor{myBlue} \cellcolor{white} \multirow{-3}{*}{\cciii} &\oursB  & 21.4 & 2.6 & 13.1 & 39.0 & 21.7 & 6.3 & 20.9 & 32.3 & 5.8 \\
    \midrule
    & \siglip-S & 31.6 & 4.8 & 29.9 & 48.3 & 38.2 & 8.6 & 28.5 & 35.0 & 10.9 \\
    \rowcolor{myBlue} \cellcolor{white} &\oursS  & 24.2 & 2.1 & 17.6 & 33.0 & 29.5 & 6.7 & 20.6 & 31.9 & 6.3 \\
    \rowcolor{myBlue} \cellcolor{white} \multirow{-3}{*}{\ccxii} &\oursB  & 30.0 & 4.2 & 26.7 & 43.3 & 37.0 & 6.4 & 22.1 & 32.4 & 5.6 \\
    \bottomrule
\end{tabular}
\label{tab:unimodal-training-res}
\end{table*}

\pagebreak
\myparagraph{Training on unimodal data.} %
As an ablation we train models on \cciii and \ccxii only, without any multimodal embedding data, corresponding to the standard \clip-like training setting. We train for 32 epochs on \cciii and for 30 epochs on \ccxii. This yields a total number of seen samples of 93M and 327M respectively, which is similar to the training runs on multimodal data as described in Sec.~\ref{sec:experiments}.
We evaluate on the same tasks as in the main paper. To this end, we use score fusion for \siglip for any task that requires multimodal embeddings at evaluation time. We do not train a \siglipmlf model in this setting, as the late fusion module would always receive a placeholder embedding for the missing modality at training time.
We report the results in Table~\ref{tab:unimodal-training-res} and observe that performance on multimodal embeddings tasks is expectedly bad when compared to multimodal training. However, on unimodal embedding tasks the performance is similar (\textit{Classification}) or improved (\textit{\imnet}).
On these tasks, \oursB is slightly better than \siglip-S when trained on \cciii, and slightly worse when trained on \ccxii.

\myparagraph{Extended MMEB results.}
We show a breakdown over all subtasks of MMEB for models trained on \cciiimm in Table~\ref{tab:res-mmeb-per-dataset-cc3m} and for models trained on \ccxiimm in Table~\ref{tab:res-mmeb-per-dataset-cc12m}.

\begin{figure*}
\centering
\footnotesize
\def\imgwidth{0.133\linewidth}
\tabcolsep=1pt
\small
\begin{tabular}{c@{\hspace{5pt}}cccccc}
\toprule
\small Query & \small \siglipSsf & \small \siglipSmlf & \small \siglipBsf & \small \siglipBmlf & \small \oursS & \small \oursB \\
\midrule
Crop to upper left & \xmark & \xmark & \xmark & \xmark & \cmark & \cmark \\
\includegraphics[width=\imgwidth]{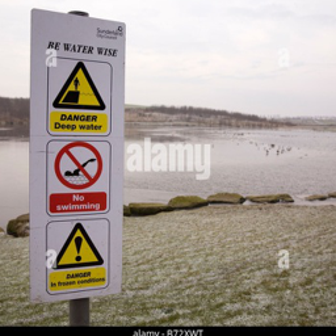} & \includegraphics[width=\imgwidth]{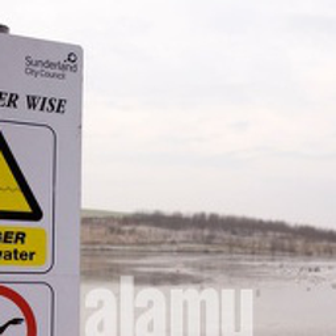} & \includegraphics[width=\imgwidth]{assets/aug-fig-neurips/randomcrop/sample0007_ret007.png} & \includegraphics[width=\imgwidth]{assets/aug-fig-neurips/randomcrop/sample0007_ret007.png} & \includegraphics[width=\imgwidth]{assets/aug-fig-neurips/randomcrop/sample0007_ret007.png} & \includegraphics[width=\imgwidth]{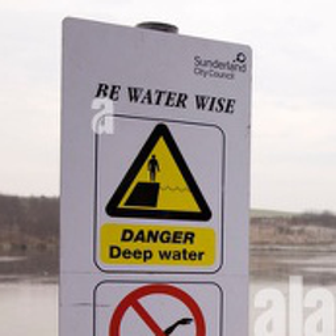} & \includegraphics[width=\imgwidth]{assets/aug-fig-neurips/randomcrop/sample0007_ret000.png} \\
\midrule
Horizontal flip & \xmark & \xmark & \xmark & \cmark & \cmark & \cmark \\
\includegraphics[width=\imgwidth]{assets/aug-fig-neurips/flip/0000_orig.png} & \includegraphics[width=\imgwidth]{assets/aug-fig-neurips/flip/sample0000_ret001.png} & \includegraphics[width=\imgwidth]{assets/aug-fig-neurips/flip/sample0000_ret001.png} & \includegraphics[width=\imgwidth]{assets/aug-fig-neurips/flip/sample0000_ret001.png} & \includegraphics[width=\imgwidth]{assets/aug-fig-neurips/flip/sample0000_ret000.png} & \includegraphics[width=\imgwidth]{assets/aug-fig-neurips/flip/sample0000_ret000.png} & \includegraphics[width=\imgwidth]{assets/aug-fig-neurips/flip/sample0000_ret000.png} \\
\midrule
\multicolumn{2}{l}{Rotate 50° clockwise \hspace{2.5mm} \xmark} & \cmark & \cmark & \cmark & \xmark & \cmark \\
\includegraphics[width=\imgwidth]{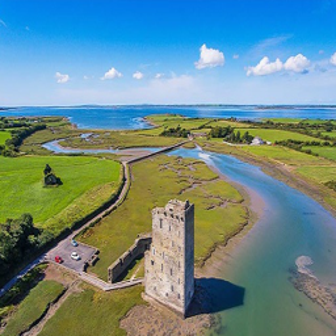} & \includegraphics[width=\imgwidth]{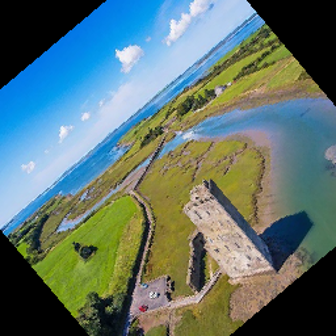} & \includegraphics[width=\imgwidth]{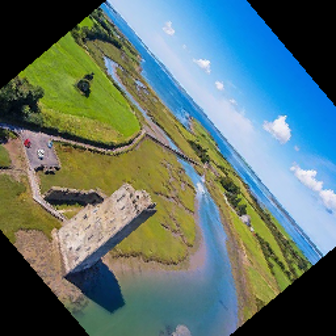} & \includegraphics[width=\imgwidth]{assets/aug-fig-neurips/rotate/sample0023_ret000.png} & \includegraphics[width=\imgwidth]{assets/aug-fig-neurips/rotate/sample0023_ret000.png} & \includegraphics[width=\imgwidth]{assets/aug-fig-neurips/rotate/sample0023_ret014.png} & \includegraphics[width=\imgwidth]{assets/aug-fig-neurips/rotate/sample0023_ret000.png} \\
\midrule
Colorize & \cmark & \cmark & \cmark & \cmark & \cmark & \cmark \\
\includegraphics[width=\imgwidth]{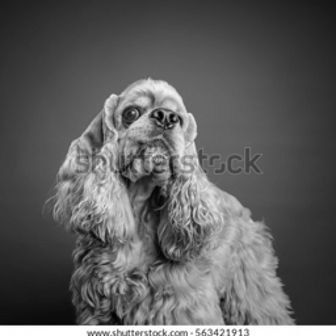} & \includegraphics[width=\imgwidth]{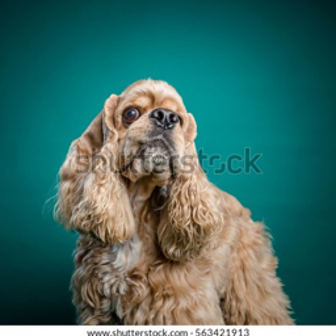} & \includegraphics[width=\imgwidth]{assets/aug-fig-neurips/colorize_grayscale/sample0028_ret000.png} & \includegraphics[width=\imgwidth]{assets/aug-fig-neurips/colorize_grayscale/sample0028_ret000.png} & \includegraphics[width=\imgwidth]{assets/aug-fig-neurips/colorize_grayscale/sample0028_ret000.png} & \includegraphics[width=\imgwidth]{assets/aug-fig-neurips/colorize_grayscale/sample0028_ret000.png} & \includegraphics[width=\imgwidth]{assets/aug-fig-neurips/colorize_grayscale/sample0028_ret000.png} \\
\midrule
\multicolumn{7}{l}{increase brightness by factor 2.0, decrease contrast by factor 0.6, increase saturation by factor 1.1} \\
& \xmark & \cmark & \xmark & \xmark & \xmark & \cmark \\
\includegraphics[width=\imgwidth]{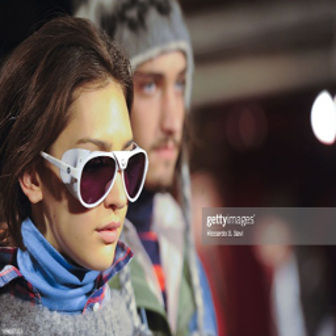} & \includegraphics[width=\imgwidth]{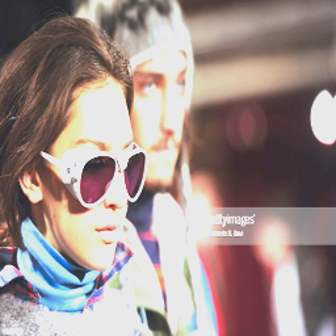} & \includegraphics[width=\imgwidth]{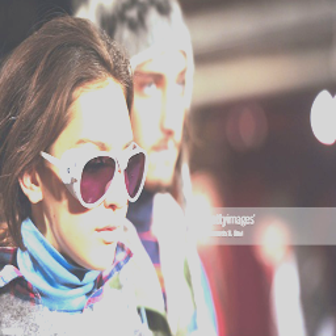} & \includegraphics[width=\imgwidth]{assets/aug-fig-neurips/colorjitterv2/sample0011_ret002.png} & \includegraphics[width=\imgwidth]{assets/aug-fig-neurips/colorjitterv2/sample0011_ret002.png} & \includegraphics[width=\imgwidth]{assets/aug-fig-neurips/colorjitterv2/sample0011_ret002.png} & \includegraphics[width=\imgwidth]{assets/aug-fig-neurips/colorjitterv2/sample0011_ret000.png} \\
\bottomrule
\end{tabular}
\caption{\textbf{\cciiiaug evaluation examples.} We illustrate the tasks in \cciiiaug, together with the prediction of the embedding models.}
\label{fig:cc3m-aug-eval}
\end{figure*}

\begin{figure*}
\centering
\footnotesize
\tabcolsep=1pt
\def\imgwidth{0.125\linewidth}
\begin{tabular}{l@{\hspace{2pt}}c@{\hspace{5pt}}cccccc}
\toprule
& \small Query & \small \siglipSsf & \small \siglipSmlf & \small \siglipBsf & \small \siglipBmlf & \small \oursS & \small \oursB \\
\midrule

\multirow{5}{*}[-45mm]{\rotatebox{90}{\small \textbf{\oicrop}}} & Bottle & \xmark & \cmark & \cmark & \cmark & \cmark & \cmark \\
& \includegraphics[width=\imgwidth]{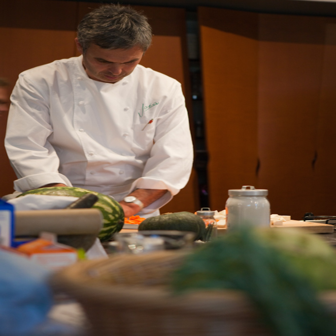} & \includegraphics[width=\imgwidth]{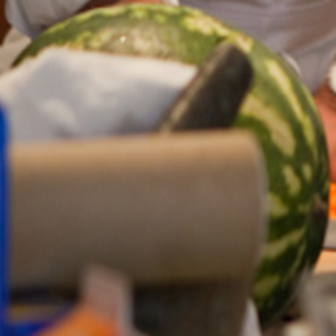} & \includegraphics[width=\imgwidth]{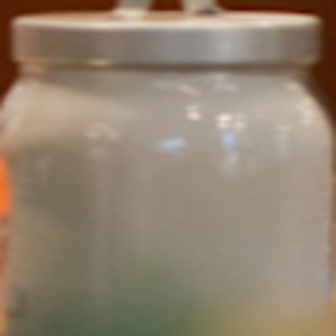} & \includegraphics[width=\imgwidth]{assets/oi-crop-fig-neurips/sample0012_ret000.png} & \includegraphics[width=\imgwidth]{assets/oi-crop-fig-neurips/sample0012_ret000.png} & \includegraphics[width=\imgwidth]{assets/oi-crop-fig-neurips/sample0012_ret000.png} & \includegraphics[width=\imgwidth]{assets/oi-crop-fig-neurips/sample0012_ret000.png} \\
\cmidrule(l){2-8}
& Footwear & \xmark & \cmark & \xmark & \cmark & \cmark & \cmark \\
& \includegraphics[width=\imgwidth]{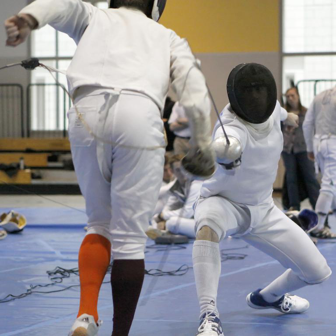} & \includegraphics[width=\imgwidth]{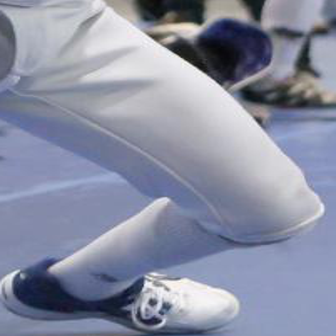} & \includegraphics[width=\imgwidth]{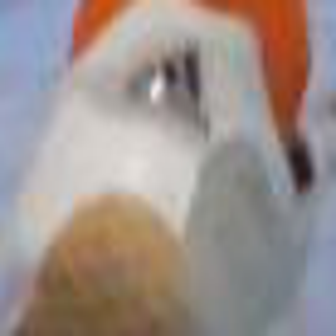} & \includegraphics[width=\imgwidth]{assets/oi-crop-fig-neurips/sample0018_ret002.png} & \includegraphics[width=\imgwidth]{assets/oi-crop-fig-neurips/sample0018_ret000.png} & \includegraphics[width=\imgwidth]{assets/oi-crop-fig-neurips/sample0018_ret000.png} & \includegraphics[width=\imgwidth]{assets/oi-crop-fig-neurips/sample0018_ret000.png} \\
\cmidrule(l){2-8}
& Racket & \xmark & \cmark & \xmark & \cmark & \cmark & \xmark \\
& \includegraphics[width=\imgwidth]{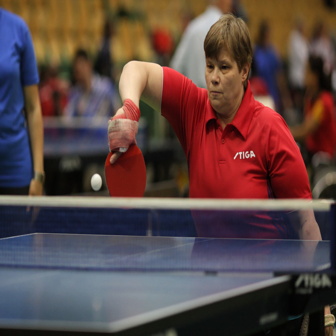} & \includegraphics[width=\imgwidth]{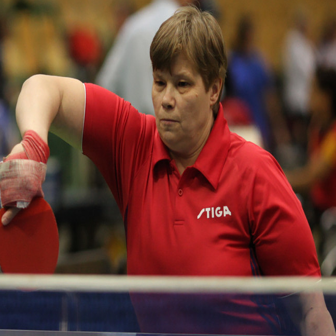} & \includegraphics[width=\imgwidth]{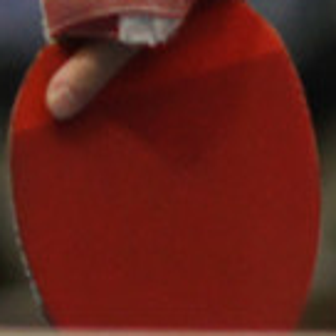} & \includegraphics[width=\imgwidth]{assets/oi-crop-fig-neurips/sample0034_ret004.png} & \includegraphics[width=\imgwidth]{assets/oi-crop-fig-neurips/sample0034_ret000.png} & \includegraphics[width=\imgwidth]{assets/oi-crop-fig-neurips/sample0034_ret000.png} & \includegraphics[width=\imgwidth]{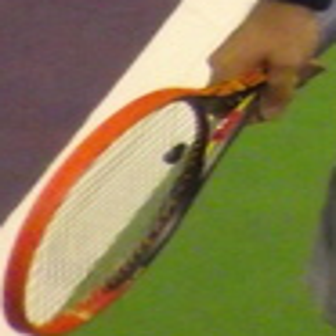} \\
\cmidrule(l){2-8}
& Picture frame & \xmark & \xmark & \xmark & \cmark & \cmark & \cmark \\
& \includegraphics[width=\imgwidth]{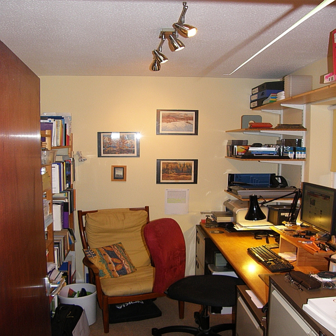} & \includegraphics[width=\imgwidth]{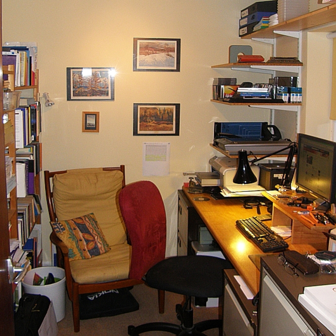} & \includegraphics[width=\imgwidth]{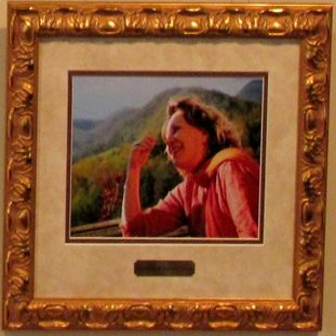} & \includegraphics[width=\imgwidth]{assets/oi-crop-fig-neurips/sample0049_ret003.png} & \includegraphics[width=\imgwidth]{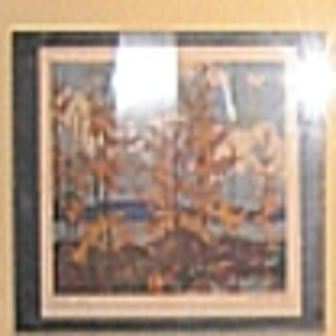} & \includegraphics[width=\imgwidth]{assets/oi-crop-fig-neurips/sample0049_ret000.png} & \includegraphics[width=\imgwidth]{assets/oi-crop-fig-neurips/sample0049_ret000.png} \\

\midrule
\midrule

\multirow{5}{*}[-45mm]{\rotatebox{90}{\small \textbf{\oipos}}} & \multicolumn{2}{l}{The cat on the right \hspace{6pt} \xmark} & \xmark & \xmark & \cmark & \cmark & \cmark \\
& \includegraphics[width=\imgwidth]{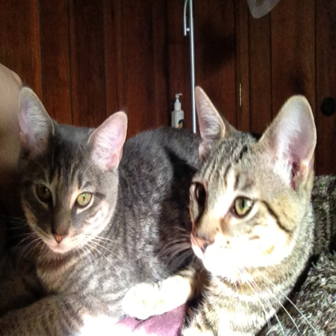} & \includegraphics[width=\imgwidth]{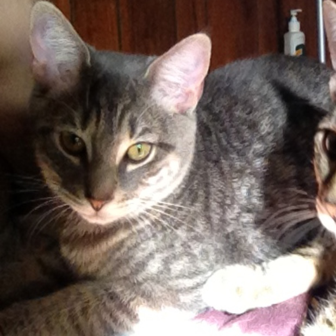} & \includegraphics[width=\imgwidth]{assets/oi-pos-fig-neurips/sample0017_ret001.png} & \includegraphics[width=\imgwidth]{assets/oi-pos-fig-neurips/sample0017_ret001.png} & \includegraphics[width=\imgwidth]{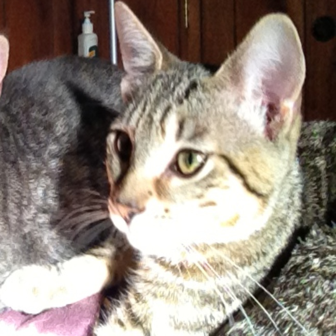} & \includegraphics[width=\imgwidth]{assets/oi-pos-fig-neurips/sample0017_ret000.png} & \includegraphics[width=\imgwidth]{assets/oi-pos-fig-neurips/sample0017_ret000.png} \\
\cmidrule(l){2-8}
\multicolumn{3}{l}{\hspace{2pt}The flower on the right \hspace{2pt} \cmark} & \xmark & \xmark & \xmark & \xmark & \cmark \\
& \includegraphics[width=\imgwidth]{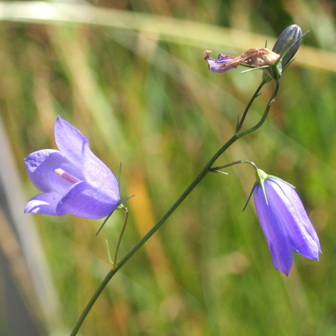} & \includegraphics[width=\imgwidth]{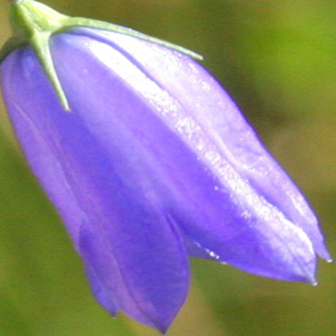} & \includegraphics[width=\imgwidth]{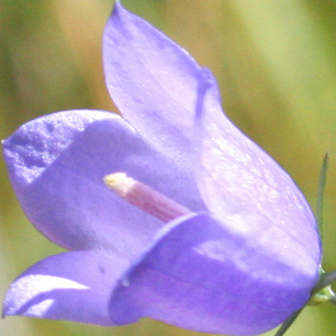} & \includegraphics[width=\imgwidth]{assets/oi-pos-fig-neurips/sample0021_ret001.png} & \includegraphics[width=\imgwidth]{assets/oi-pos-fig-neurips/sample0021_ret001.png} & \includegraphics[width=\imgwidth]{assets/oi-pos-fig-neurips/sample0021_ret001.png} & \includegraphics[width=\imgwidth]{assets/oi-pos-fig-neurips/sample0021_ret000.png} \\
\cmidrule(l){2-8}
& \multicolumn{2}{l}{The dog on the left \hspace{6pt} \cmark} & \cmark & \cmark & \cmark & \cmark & \cmark \\
& \includegraphics[width=\imgwidth]{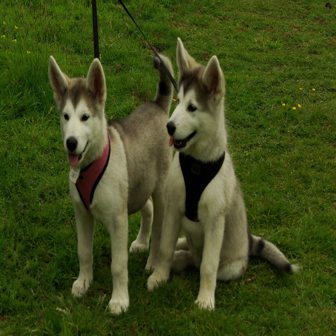} & \includegraphics[width=\imgwidth]{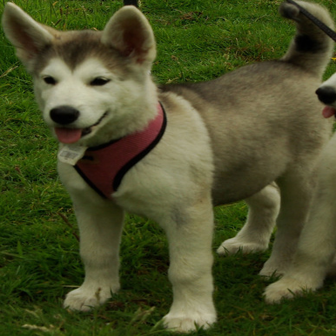} & \includegraphics[width=\imgwidth]{assets/oi-pos-fig-neurips/sample0000_ret000.png} & \includegraphics[width=\imgwidth]{assets/oi-pos-fig-neurips/sample0000_ret000.png} & \includegraphics[width=\imgwidth]{assets/oi-pos-fig-neurips/sample0000_ret000.png} & \includegraphics[width=\imgwidth]{assets/oi-pos-fig-neurips/sample0000_ret000.png} & \includegraphics[width=\imgwidth]{assets/oi-pos-fig-neurips/sample0000_ret000.png} \\
\cmidrule(l){2-8}
& \multicolumn{2}{l}{The horse on the left \hspace{3pt} \cmark} & \xmark & \cmark & \xmark & \cmark & \cmark \\
& \includegraphics[width=\imgwidth]{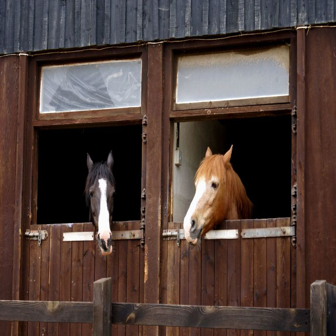} & \includegraphics[width=\imgwidth]{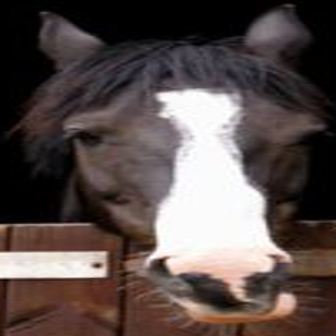} & \includegraphics[width=\imgwidth]{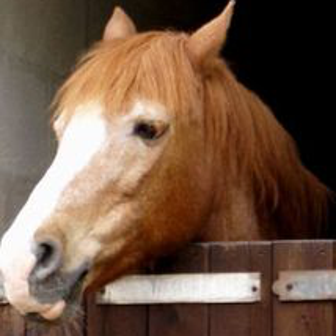} & \includegraphics[width=\imgwidth]{assets/oi-pos-fig-neurips/sample0042_ret000.png} & \includegraphics[width=\imgwidth]{assets/oi-pos-fig-neurips/sample0042_ret001.png} & \includegraphics[width=\imgwidth]{assets/oi-pos-fig-neurips/sample0042_ret000.png} & \includegraphics[width=\imgwidth]{assets/oi-pos-fig-neurips/sample0042_ret000.png} \\

\bottomrule
\end{tabular}
\caption{\textbf{\oicrop and \oipos.} We show images retrieved on the \oicrop and \oipos tasks by models trained on \ccxiimm.}
\label{fig:oi-crop-pos-eval}
\end{figure*}

\begin{figure*}
\centering
\footnotesize
\tabcolsep=1pt
\def\imgwidth{0.135\linewidth}
\begin{tabular}{c@{\hspace{5pt}}cccccc}
\toprule
\small Query & \small \siglipSsf & \small \siglipSmlf & \small \siglipBsf & \small \siglipBmlf & \small \oursS & \small \oursB \\
\midrule
\multicolumn{6}{l}{First elephant is up the hill }\\ & \xmark & \cmark & \xmark & \xmark & \xmark & \cmark \\
\includegraphics[width=\imgwidth]{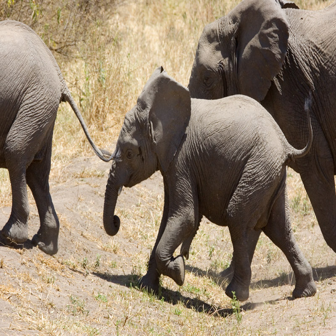} & \includegraphics[width=\imgwidth]{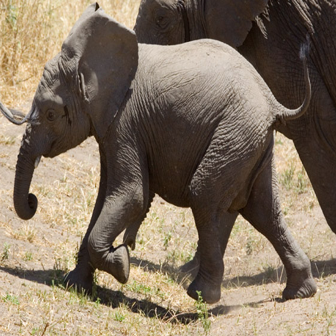} & \includegraphics[width=\imgwidth]{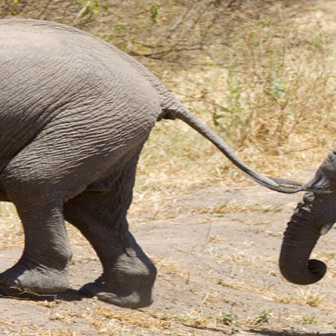} & \includegraphics[width=\imgwidth]{assets/vg-crop-fig-neurips/sample0045_ret043.png} & \includegraphics[width=\imgwidth]{assets/vg-crop-fig-neurips/sample0045_ret043.png} & \includegraphics[width=\imgwidth]{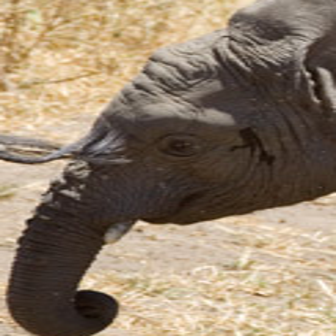} & \includegraphics[width=\imgwidth]{assets/vg-crop-fig-neurips/sample0045_ret045.png} \\
\midrule
\multicolumn{6}{l}{stapler on the stand}\\ & \xmark & \xmark & \xmark & \xmark & \xmark & \cmark \\
\includegraphics[width=\imgwidth]{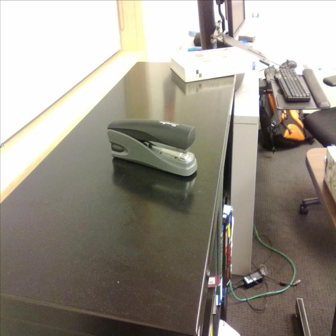} & \includegraphics[width=\imgwidth]{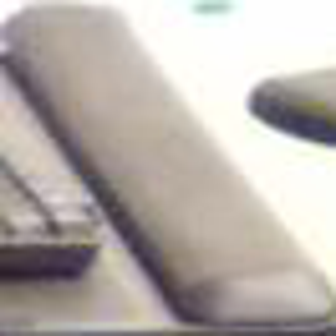} & \includegraphics[width=\imgwidth]{assets/vg-crop-fig-neurips/sample0083_ret094.png} & \includegraphics[width=\imgwidth]{assets/vg-crop-fig-neurips/sample0083_ret094.png} & \includegraphics[width=\imgwidth]{assets/vg-crop-fig-neurips/sample0083_ret094.png} & \includegraphics[width=\imgwidth]{assets/vg-crop-fig-neurips/sample0083_ret094.png} & \includegraphics[width=\imgwidth]{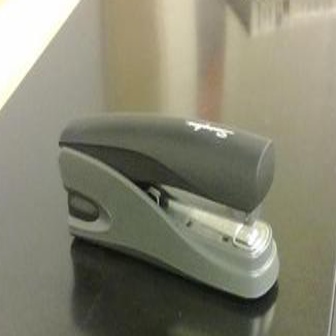} \\
\midrule
\multicolumn{6}{l}{A fork on a white plate}\\ & \xmark & \cmark & \cmark & \cmark & \cmark & \xmark \\
\includegraphics[width=\imgwidth]{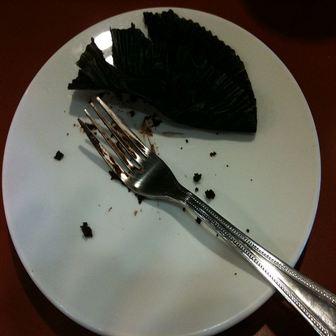} & \includegraphics[width=\imgwidth]{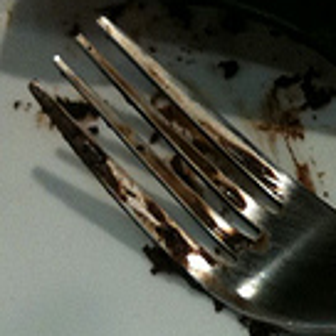} & \includegraphics[width=\imgwidth]{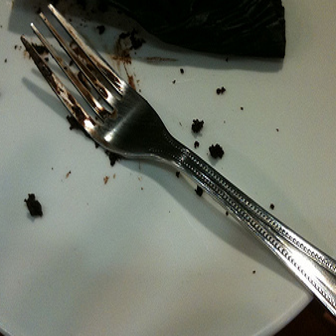} & \includegraphics[width=\imgwidth]{assets/vg-crop-fig-neurips/sample0358_ret358.png} & \includegraphics[width=\imgwidth]{assets/vg-crop-fig-neurips/sample0358_ret358.png} & \includegraphics[width=\imgwidth]{assets/vg-crop-fig-neurips/sample0358_ret358.png} & \includegraphics[width=\imgwidth]{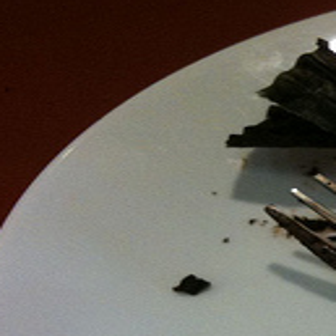} \\
\midrule
\multicolumn{6}{l}{the sign is yellow in color}\\ & \cmark & \cmark & \cmark & \cmark & \cmark & \cmark \\
\includegraphics[width=\imgwidth]{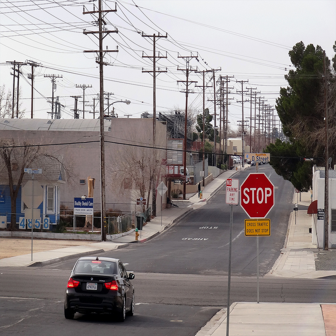} & \includegraphics[width=\imgwidth]{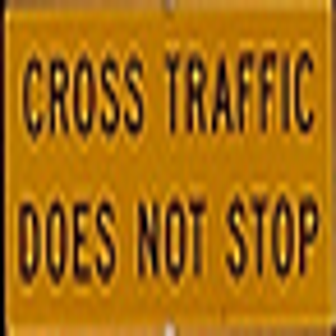} & \includegraphics[width=\imgwidth]{assets/vg-crop-fig-neurips/sample0215_ret215.png} & \includegraphics[width=\imgwidth]{assets/vg-crop-fig-neurips/sample0215_ret215.png} & \includegraphics[width=\imgwidth]{assets/vg-crop-fig-neurips/sample0215_ret215.png} & \includegraphics[width=\imgwidth]{assets/vg-crop-fig-neurips/sample0215_ret215.png} & \includegraphics[width=\imgwidth]{assets/vg-crop-fig-neurips/sample0215_ret215.png} \\
\midrule
\multicolumn{6}{l}{Two people on sidewalk}\\ & \cmark & \xmark & \cmark & \xmark & \cmark & \cmark \\
\includegraphics[width=\imgwidth]{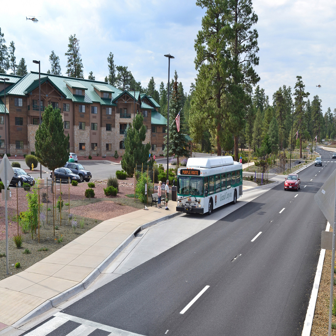} & \includegraphics[width=\imgwidth]{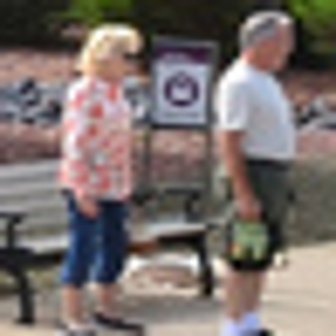} & \includegraphics[width=\imgwidth]{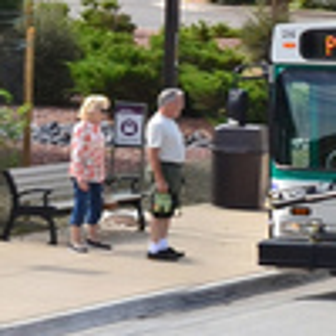} & \includegraphics[width=\imgwidth]{assets/vg-crop-fig-neurips/sample0252_ret252.png} & \includegraphics[width=\imgwidth]{assets/vg-crop-fig-neurips/sample0252_ret261.png} & \includegraphics[width=\imgwidth]{assets/vg-crop-fig-neurips/sample0252_ret252.png} & \includegraphics[width=\imgwidth]{assets/vg-crop-fig-neurips/sample0252_ret252.png} \\
\midrule
\multicolumn{6}{l}{Red car on road}\\ & \cmark & \cmark & \cmark & \cmark & \cmark & \cmark \\
\includegraphics[width=\imgwidth]{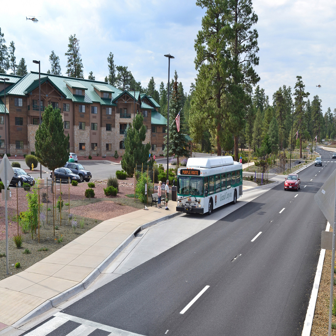} & \includegraphics[width=\imgwidth]{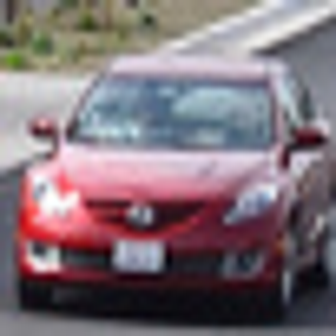} & \includegraphics[width=\imgwidth]{assets/vg-crop-fig-neurips/sample0253_ret253.png} & \includegraphics[width=\imgwidth]{assets/vg-crop-fig-neurips/sample0253_ret253.png} & \includegraphics[width=\imgwidth]{assets/vg-crop-fig-neurips/sample0253_ret253.png} & \includegraphics[width=\imgwidth]{assets/vg-crop-fig-neurips/sample0253_ret253.png} & \includegraphics[width=\imgwidth]{assets/vg-crop-fig-neurips/sample0253_ret253.png} \\
\bottomrule
\end{tabular}
\caption{\textbf{\vgcrop.} We show images retrieved on the \vgcrop task by models trained on \ccxiimm.}
\label{fig:vg-crop-eval}
\end{figure*}

\myparagraph{Qualitative examples.}
We show qualitative examples of images retrieved by models trained on \ccxiimm on \cciiiaug in Fig.~\ref{fig:cc3m-aug-eval}, \oicrop and \oipos in Fig.~\ref{fig:oi-crop-pos-eval}, and \vgcrop in Fig.~\ref{fig:vg-crop-eval}.

\begin{table*}[t]
\small
\caption{%
\textbf{Contribution of the MMM loss.} Results on additional pre-training data.
}
\centering
\tabcolsep=1.5pt
\begin{tabular}{l l ccccccccc
}
    \toprule
    Train Data & Model & $\L_\textrm{MMM}$ & Classification\zs & VQA\zs & Retrieval\zs & Grounding\zs & ImNet\zs & VG-Crop & OI-Crop\zs & TGIT \\
    \midrule

    & \oursS & \xmark  & 16.3 & 0.8 & 7.6 & 23.3 & 14.3 & 6.5 & 16.2 & 8.4 \\
    & \oursS & \cmark & 16.7 & 1.1 & 8.1 & 22.0 & 15.4 & 6.5 & 19.2 & 7.6 \\
    & \oursB & \xmark & 18.8 & 1.8 & 12.6 & 36.2 & 19.6 & 6.2 & 20.6 & 5.6 \\
    \multirow{-4}{*}{\cciii} &\oursB & \cmark & 21.4 & 2.6 & 13.1 & 39.0 & 21.7 & 6.3 & 20.9 & 5.8 \\
    \midrule

    & \oursS & \xmark & 23.1 & 2.0 & 16.9 & 35.4 & 29.0 & 6.3 & 20.8 & 5.9 \\
    &\oursS &\cmark & 24.2 & 2.1 & 17.6 & 33.0 & 29.5 & 6.7 & 20.6 & 6.3 \\
    & \oursB & \xmark & 26.8 & 3.5 & 22.0 & 40.1 & 35.0 & 6.4 & 20.5 & 5.9 \\
    \multirow{-4}{*}{\ccxii} &\oursB & \cmark & 30.0 & 4.2 & 26.7 & 43.3 & 37.0 & 6.4 & 22.1 & 5.6 \\
    \midrule

    & \oursS & \xmark & 23.5 & 17.7 & 19.2 & 69.9 & 24.4 & 53.3 & 63.1 & 89.3 \\
    & \oursS & \cmark & 25.2 & 18.2 & 20.1 & 75.2 & 26.0 & 53.5 & 64.7 & 90.6 \\
    & \oursB & \xmark & 28.3 & 19.1 & 23.8 & 81.8 & 30.8 & 59.5 & 70.6 & 95.8 \\
    \multirow{-4}{*}{\ccxiimm} & \oursB & \cmark & 31.2 & 19.8 & 26.2 & 82.3 & 32.7 & 61.5 & 71.3 & 94.2 \\
    \bottomrule
\end{tabular}
\label{tab:ablation_mlm_extended}
\end{table*}

\begin{table*}
\centering
\caption{\textbf{MMEB breakdown} for models trained on \cciiimm.}
\label{tab:res-mmeb-per-dataset-cc3m}
\tabcolsep=3pt
\small
\begin{tabular}{llccccaa}
\toprule
 && \siglipSsf & \siglipSmlf & \siglipBsf & \siglipBmlf & \oursS & \oursB \\
\midrule
& ImageNet-1K & 10.7 & 11.1 & 14.2 & 13.8 & 12.3 & 17.4 \\
& CIFAR-100 & 16.7 & 16.2 & 18.1 & 20.4 & 13.8 & 26.1 \\
& N24News & 27.9 & 23.4 & 28.7 & 21.2 & 24.2 & 25.5 \\
& HatefulMemes & 51.3 & 51.0 & 51.8 & 51.0 & 50.6 & 52.2 \\
& VOC2007 & 70.2 & 40.4 & 69.5 & 39.9 & 43.0 & 43.6 \\
& SUN397 & 24.0 & 25.5 & 25.4 & 29.3 & 28.1 & 38.5 \\
& Place365 & 18.8 & 15.0 & 17.6 & 17.8 & 17.2 & 23.4 \\
& ImageNet-A & 0.6 & 0.9 & 0.8 & 1.3 & 0.4 & 0.6 \\
& ImageNet-R & 11.5 & 8.7 & 12.7 & 12.7 & 6.8 & 20.0 \\
& ObjectNet & 4.2 & 5.0 & 4.8 & 6.1 & 6.1 & 7.7 \\
& Country211 & 0.4 & 1.2 & 0.4 & 1.1 & 1.2 & 1.4 \\
\cmidrule{2-8}
\multirow{-13}{*}{\rotatebox{90}{\textbf{Classification}}} & \textbf{Mean} & 21.5 & 18.0 & 22.2 & 19.5 & 18.5 & 23.3 \\
\midrule
& OK-VQA & 20.8 & 19.8 & 20.7 & 21.8 & 23.6 & 26.3 \\
& A-OKVQA & 18.1 & 17.8 & 19.0 & 19.9 & 18.6 & 22.8 \\
& DocVQA & 2.6 & 3.4 & 2.8 & 3.5 & 3.7 & 4.5 \\
& InfographicsVQA & 2.0 & 3.6 & 2.7 & 2.7 & 3.6 & 4.1 \\
& ChartQA & 5.4 & 3.2 & 4.4 & 2.6 & 4.0 & 4.2 \\
& Visual7W & 22.3 & 23.8 & 25.0 & 25.3 & 24.6 & 29.7 \\
& ScienceQA & 3.0 & 4.2 & 4.6 & 4.4 & 4.2 & 4.4 \\
& VizWiz & 3.7 & 4.2 & 3.8 & 3.9 & 5.2 & 6.0 \\
& GQA & 42.5 & 55.0 & 46.6 & 57.3 & 65.3 & 66.4 \\
& TextVQA & 6.7 & 6.5 & 6.1 & 6.3 & 6.1 & 6.7 \\
\cmidrule{2-8}
\multirow{-12}{*}{\rotatebox{90}{\textbf{VQA}}} & \textbf{Mean} & 12.7 & 14.2 & 13.6 & 14.8 & 15.9 & 17.5 \\
\midrule
& VisDial & 6.4 & 6.0 & 7.1 & 7.2 & 1.9 & 0.9 \\
& CIRR & 13.4 & 13.8 & 13.2 & 14.5 & 10.2 & 15.3 \\
& VisualNews-t2i & 9.3 & 8.6 & 10.9 & 8.5 & 7.2 & 9.1 \\
& VisualNews-i2t & 10.4 & 9.2 & 11.7 & 10.6 & 7.7 & 11.7 \\
& MSCOCO-t2i & 23.3 & 25.5 & 24.5 & 29.0 & 23.1 & 32.3 \\
& MSCOCO-i2t & 20.1 & 21.4 & 20.8 & 24.7 & 21.3 & 31.0 \\
& NIGHTS & 44.2 & 42.7 & 42.9 & 45.6 & 38.2 & 46.0 \\
& WebQA & 8.4 & 10.0 & 8.5 & 10.0 & 13.5 & 15.6 \\
& FashionIQ & 5.7 & 4.6 & 7.1 & 4.3 & 2.2 & 5.3 \\
& Wiki-SS-NQ & 1.0 & 0.9 & 0.9 & 1.1 & 0.9 & 1.1 \\
& OVEN & 7.3 & 5.2 & 6.4 & 6.9 & 4.5 & 6.4 \\
& EDIS & 6.7 & 4.7 & 6.6 & 3.9 & 4.1 & 5.9 \\
\cmidrule{2-8}
\multirow{-14}{*}{\rotatebox{90}{\textbf{Retrieval}}} & \textbf{Mean} & 13.0 & 12.7 & 13.4 & 13.9 & 11.2 & 15.0 \\
\midrule
& MSCOCO & 69.7 & 73.9 & 72.1 & 73.9 & 64.4 & 78.9 \\
& RefCOCO & 81.8 & 82.8 & 83.5 & 85.5 & 76.4 & 91.3 \\
& RefCOCO-Matching & 60.1 & 63.1 & 61.8 & 63.1 & 65.0 & 68.8 \\
& Visual7W-Pointing & 87.7 & 77.0 & 91.3 & 85.0 & 77.3 & 90.8 \\
\cmidrule{2-8}
\multirow{-5}{*}[1pt]{\rotatebox{90}{\textbf{Grounding}}} & \textbf{Mean} & 74.8 & 74.2 & 77.2 & 76.9 & 70.8 & 82.4 \\
\bottomrule
\end{tabular}
\end{table*}

\begin{table*}
\centering
\caption{\textbf{MMEB breakdown} for models trained on \ccxiimm.}
\label{tab:res-mmeb-per-dataset-cc12m}
\tabcolsep=3pt
\small
\begin{tabular}{llccccaa}
\toprule
 && \siglipSsf & \siglipSmlf & \siglipBsf & \siglipBmlf & \oursS & \oursB \\
\midrule
& ImageNet-1K & 25.8 & 29.6 & 28.7 & 29.7 & 25.0 & 31.2 \\
& CIFAR-100 & 23.7 & 29.1 & 28.0 & 29.7 & 21.1 & 35.7 \\
& N24News & 34.9 & 15.7 & 35.2 & 27.3 & 24.8 & 29.1 \\
& HatefulMemes & 50.0 & 51.0 & 48.8 & 52.3 & 48.6 & 50.7 \\
& VOC2007 & 78.6 & 56.6 & 77.9 & 60.9 & 52.7 & 55.8 \\
& SUN397 & 43.6 & 46.4 & 45.1 & 49.3 & 41.4 & 50.5 \\
& Place365 & 30.5 & 29.3 & 28.3 & 26.6 & 28.4 & 32.2 \\
& ImageNet-A & 1.2 & 1.0 & 1.7 & 2.0 & 0.9 & 1.3 \\
& ImageNet-R & 31.7 & 38.7 & 37.4 & 36.4 & 21.5 & 38.5 \\
& ObjectNet & 10.4 & 11.7 & 11.7 & 14.3 & 9.2 & 13.5 \\
& Country211 & 4.3 & 4.5 & 3.9 & 4.4 & 3.6 & 4.6 \\
\cmidrule{2-8}
\multirow{-13}{*}{\rotatebox{90}{\textbf{Classification}}} & \textbf{Mean} & 30.4 & 28.5 & 31.5 & 30.3 & 25.2 & 31.2 \\
\midrule
& OK-VQA & 28.7 & 29.1 & 30.3 & 28.0 & 28.8 & 30.9 \\
& A-OKVQA & 26.3 & 24.2 & 26.0 & 25.5 & 24.4 & 28.2 \\
& DocVQA & 3.0 & 3.5 & 3.1 & 3.5 & 5.9 & 4.2 \\
& InfographicsVQA & 2.8 & 3.4 & 3.9 & 4.2 & 4.4 & 4.9 \\
& ChartQA & 1.7 & 2.7 & 2.5 & 1.9 & 3.0 & 2.9 \\
& Visual7W & 30.0 & 30.2 & 29.5 & 32.1 & 31.3 & 35.3 \\
& ScienceQA & 3.4 & 4.8 & 4.2 & 2.9 & 4.7 & 5.4 \\
& VizWiz & 5.3 & 6.9 & 6.0 & 6.9 & 6.6 & 8.7 \\
& GQA & 52.3 & 55.0 & 54.2 & 54.7 & 63.9 & 68.0 \\
& TextVQA & 8.2 & 8.8 & 9.9 & 8.4 & 9.2 & 9.2 \\
\cmidrule{2-8}
\multirow{-12}{*}{\rotatebox{90}{\textbf{VQA}}} & \textbf{Mean} & 16.2 & 16.9 & 17.0 & 16.8 & 18.2 & 19.8 \\
\midrule
& VisDial & 9.2 & 7.4 & 11.6 & 6.4 & 10.3 & 13.9 \\
& CIRR & 22.6 & 22.5 & 22.0 & 23.5 & 15.1 & 22.5 \\
& VisualNews-t2i & 26.2 & 25.5 & 28.2 & 27.9 & 21.5 & 27.5 \\
& VisualNews-i2t & 25.0 & 24.4 & 26.5 & 26.7 & 21.8 & 27.0 \\
& MSCOCO-t2i & 43.4 & 42.3 & 44.3 & 41.4 & 36.5 & 46.2 \\
& MSCOCO-i2t & 38.3 & 40.9 & 41.1 & 40.7 & 36.8 & 46.2 \\
& NIGHTS & 56.8 & 58.9 & 47.6 & 49.7 & 49.1 & 57.7 \\
& WebQA & 21.3 & 22.0 & 22.7 & 25.0 & 22.8 & 24.6 \\
& FashionIQ & 9.0 & 4.5 & 10.6 & 5.8 & 4.8 & 8.2 \\
& Wiki-SS-NQ & 2.9 & 2.9 & 4.0 & 3.9 & 1.4 & 3.2 \\
& OVEN & 12.0 & 13.2 & 9.8 & 13.6 & 7.9 & 16.8 \\
& EDIS & 18.9 & 14.2 & 17.0 & 13.2 & 13.6 & 21.1 \\
\cmidrule{2-8}
\multirow{-14}{*}{\rotatebox{90}{\textbf{Retrieval}}} & \textbf{Mean} & 23.8 & 23.2 & 23.8 & 23.2 & 20.1 & 26.2 \\
\midrule
& MSCOCO & 72.3 & 73.7 & 73.5 & 78.0 & 71.8 & 80.9 \\
& RefCOCO & 78.5 & 81.5 & 78.6 & 83.4 & 82.6 & 91.7 \\
& RefCOCO-Matching & 59.3 & 62.2 & 57.2 & 61.9 & 71.5 & 71.5 \\
& Visual7W-Pointing & 86.5 & 73.3 & 81.4 & 70.3 & 74.8 & 85.0 \\
\cmidrule{2-8}
\multirow{-5}{*}[1pt]{\rotatebox{90}{\textbf{Grounding}}} & \textbf{Mean} & 74.2 & 72.7 & 72.7 & 73.4 & 75.2 & 82.3 \\
\bottomrule
\end{tabular}
\end{table*}

\end{document}